\documentclass[10pt,twocolumn,letterpaper]{article}

\usepackage{iccv}
\usepackage{times}
\usepackage{epsfig}
\usepackage{graphicx}
\usepackage{amsmath}
\usepackage{amssymb}

\usepackage{longtable}

\usepackage[breaklinks=true,bookmarks=false]{hyperref}

\begin{document}
\pagestyle{plain}

%%%%%%%%% TITLE
\title{TextureSAM: Towards a Texture Aware Foundation Model for Segmentation}

\author{
Inbal Cohen$^{\dagger}$\\
Tel Aviv University, Israel\\
{\tt\small inbalc2@mail.tau.ac.il}
\and
Boaz Meivar$^{\dagger}$\\
Tel Aviv University, Israel\\
{\tt\small boazmeivar@mail.tau.ac.il}
\and
Peihan Tu\\
University of Maryland, College Park, USA\\
{\tt\small phtu@terpmail.umd.edu}
\and
Shai Avidan\\
Tel Aviv University, Israel\\
{\tt\small avidan@eng.tau.ac.il}
\and
Gal Oren\\
Stanford University, Technion, USA\\
{\tt\small galoren@stanford.edu}
}

%\thispagestyle{empty}

%% dataset

\newcommand{\appfeature}{f}
\newcommand{\gaussian}{g}
\newcommand{\gaussiancenter}{\mu}
\newcommand{\gaussiancovariance}{\Sigma}
\newcommand{\gaussianindex}{i}
\newcommand{\gaussianindexed}{\gaussian_{\gaussianindex}}

\newcommand{\texturesymbol}{t}
\newcommand{\texture}{\texturesymbol}

\newcommand{\appearancefeature}{f}

\newcommand{\setof}[1]{\{#1\}}
\newcommand{\gaussianset}{\mathcal{G}}

\newcommand{\encoder}{E}
\newcommand{\decoder}{D}

\newcommand{\imagesymbol}{I}
\newcommand{\imagesize}{l}
\newcommand{\contentsymbol}{c}

\newcommand{\textureimage}{\imagesymbol_\texturesymbol}
\newcommand{\textureimageset}{\mathcal{T}}

\newcommand{\contentimage}{\imagesymbol_\contentsymbol}

\newcommand{\patchsymbol}{p}
\newcommand{\patch}{\patchsymbol}
\newcommand{\patchset}{\mathcal{P}}
\newcommand{\numpatches}{\#patches}

\newcommand{\contentimagepatch}{\contentimage^\patchsymbol}

\newcommand{\gaussianpatch}{\gaussian^{\patchsymbol}}
\newcommand{\gaussiantexture}{\gaussian_{\texturesymbol}}
\newcommand{\concatenate}{\text{Concat}}

\newcommand{\gaussiansetcontent}{\gaussianset_\contentsymbol}
\newcommand{\gaussiansetcontentpatch}{\gaussianset_\contentsymbol^\patchsymbol}
\newcommand{\gaussiansettexture}{\gaussianset_\texturesymbol}

\newcommand{\masksymbol}{m}
\newcommand{\mask}{\masksymbol}
\newcommand{\maskset}{\mathcal{M}}

\newcommand{\modified}[1]{\tilde{#1}}

\newcommand{\contentmasked}[1]{{#1}_\contentsymbol^\masksymbol}
\newcommand{\gaussiancontent}{\gaussian_\contentsymbol}

\newcommand{\interpcoefficient}{\eta}
\newcommand{\realnumberset}{\mathbb{R}}
\newcommand{\assign}{\leftarrow}

\newcommand{\merge}{\mathsf{Merge}}
\newcommand{\scale}{\mathsf{Scale}}
\newcommand{\size}{\mathsf{Size}}
\newcommand{\sizeof}[1]{S_{#1}}
%\begin{document}
\twocolumn[{%
\renewcommand\twocolumn[1][]{#1}%
\maketitle
\begin{center}
    \centering
    \captionsetup{type=figure}
    \begin{tabular}{c c c c c c c c }
        {\raisebox{3\height}{\textbf{Image}}} &  
        \includegraphics[width=0.1\linewidth]{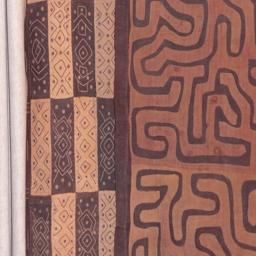} &
        \includegraphics[width=0.1\linewidth]{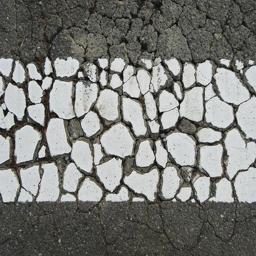} &
        \includegraphics[width=0.1\linewidth]{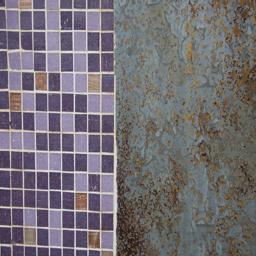} &
        \includegraphics[width=0.1\linewidth]{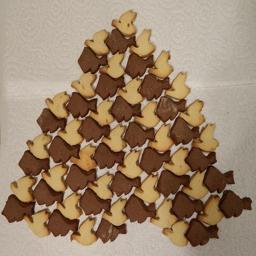} &
        \includegraphics[width=0.1\linewidth]{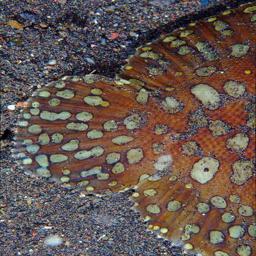} &
        \includegraphics[width=0.1\linewidth]{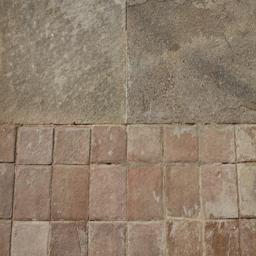} &        
        \includegraphics[width=0.1\linewidth]{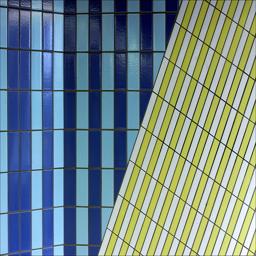} \\

        {\raisebox{3\height}{\textbf{SAM-2}}} &  
        \includegraphics[width=0.1\linewidth]{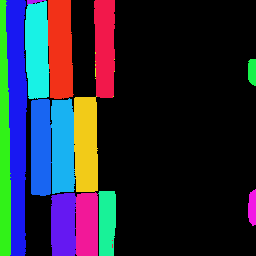} &
        \includegraphics[width=0.1\linewidth]{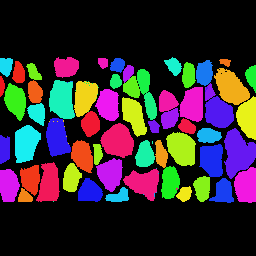} &
        \includegraphics[width=0.1\linewidth]{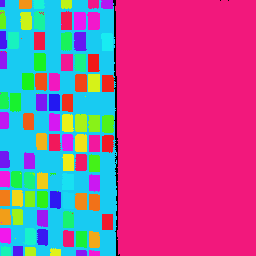} &
        \includegraphics[width=0.1\linewidth]{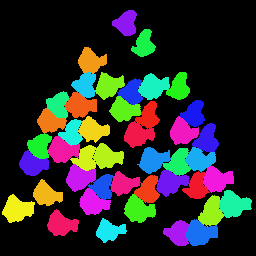} &
        \includegraphics[width=0.1\linewidth]{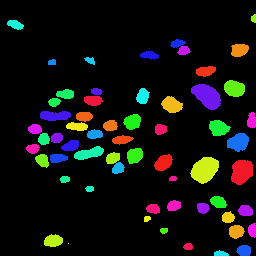} &
        \includegraphics[width=0.1\linewidth]{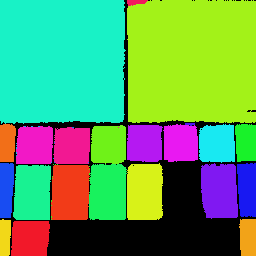} &             
        \includegraphics[width=0.1\linewidth]{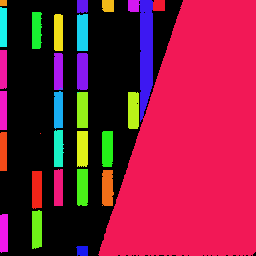} \\

        {\raisebox{3\height}{\textbf{SAM-2*}}} &  
        \includegraphics[width=0.1\linewidth]{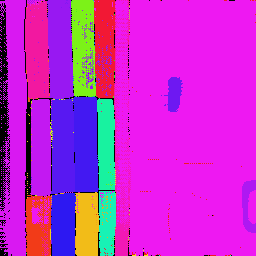} &
        \includegraphics[width=0.1\linewidth]{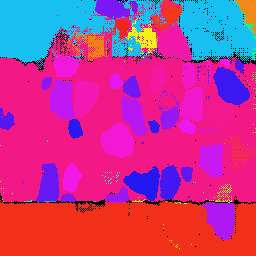} &
        \includegraphics[width=0.1\linewidth]{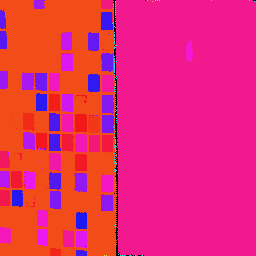} &
        \includegraphics[width=0.1\linewidth]{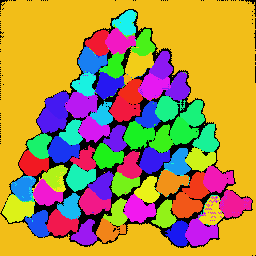} &
        \includegraphics[width=0.1\linewidth]{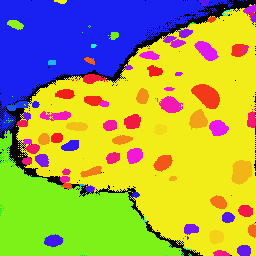} &
        \includegraphics[width=0.1\linewidth]{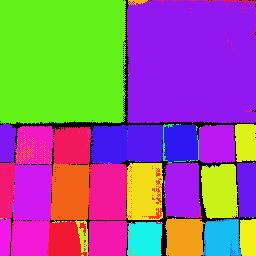} &             
        \includegraphics[width=0.1\linewidth]{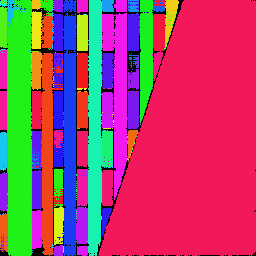} \\

        {\raisebox{3\height}{\textbf{TextureSAM}}} &  
        \includegraphics[width=0.1\linewidth]{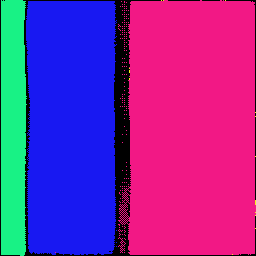} &
        \includegraphics[width=0.1\linewidth]{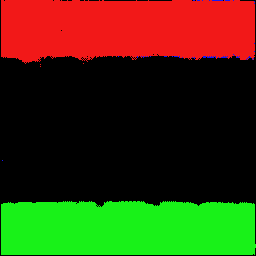} &
        \includegraphics[width=0.1\linewidth]{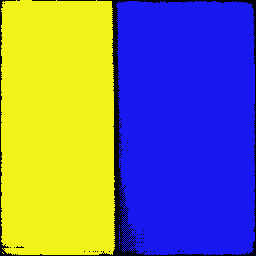} &
        \includegraphics[width=0.1\linewidth]{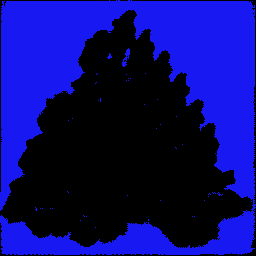} &
        \includegraphics[width=0.1\linewidth]{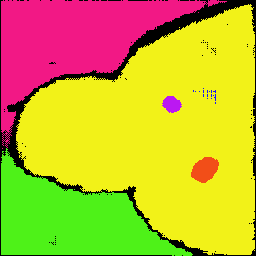} &
        \includegraphics[width=0.1\linewidth]{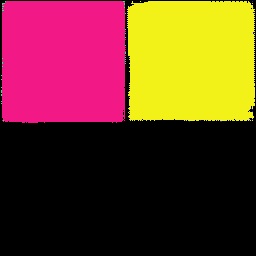} &   
        \includegraphics[width=0.1\linewidth]{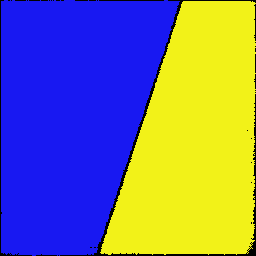} \\

        {\raisebox{3\height}{\textbf{GT}}} &  
        \includegraphics[width=0.1\linewidth]{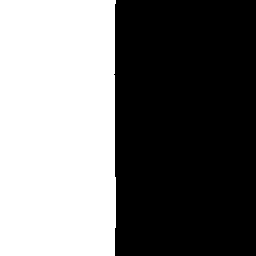} &
        \includegraphics[width=0.1\linewidth]{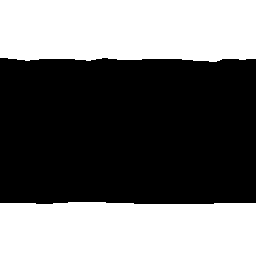} &
        \includegraphics[width=0.1\linewidth]{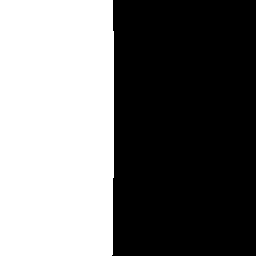} &
        \includegraphics[width=0.1\linewidth]{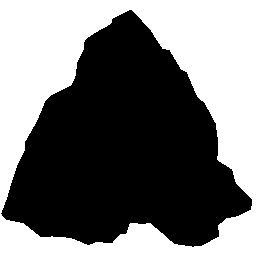} &
        \includegraphics[width=0.1\linewidth]{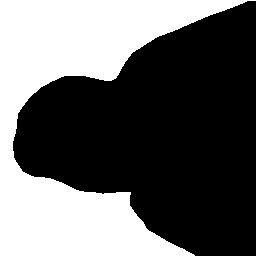} &
        \includegraphics[width=0.1\linewidth]{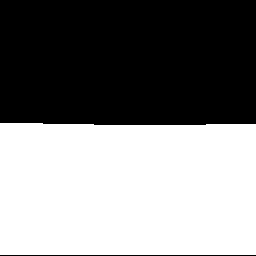} &  
        \includegraphics[width=0.1\linewidth]{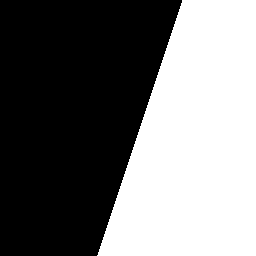} \\        
    \end{tabular}
    \caption{Examples for segmentation of natural images from the Real-World Textured Dataset. Compared with the original SAM (2nd and 3rd rows, SAM-2* indicating modified inference parameters), TextureSAM (4th row) is more adept at recognizing areas defined by texture changes. SAM's reliance on semantic shape results in fragmented segmentation, while TextureSAM provides segmentation of whole areas defined by texture changes. }
    \label{fig:teaser}
\end{center}
}]

\let\thefootnote\relax\footnotetext{$^{\dagger}$Equal contribution.}

\begin{abstract}

Segment Anything Models (SAM) have achieved remarkable success in object segmentation tasks across diverse datasets. However, these models are predominantly trained on large-scale semantic segmentation datasets, which introduce a bias toward object shape rather than texture cues in the image. This limitation is critical in domains such as medical imaging, material classification, and remote sensing, where texture changes define object boundaries. In this study, we investigate SAM’s bias toward semantics over textures and introduce a new texture-aware foundation model, TextureSAM, which performs superior segmentation in texture-dominant scenarios. To achieve this, we employ a novel fine-tuning approach that incorporates texture augmentation techniques, incrementally modifying training images to emphasize texture features. By leveraging a novel texture-alternation of the ADE20K dataset, we guide TextureSAM to prioritize texture-defined regions, thereby mitigating the inherent shape bias present in the original SAM model. Our extensive experiments demonstrate that TextureSAM significantly outperforms SAM-2 on both natural (\textbf{+0.2 mIoU}) and synthetic (\textbf{+0.18 mIoU}) texture-based segmentation datasets. The code and texture-augmented dataset will be publicly available.
\end{abstract}

\section{Introduction}

Conventional segmentation models are designed to recognize objects based on semantic features, yet many real-world applications rely on texture, which is inherently difficult to define. Texture can manifest as structured, repeating patterns or irregular, stochastic variations, making it challenging to model in a unified framework. In biomedical imaging~\cite{castellano2004texture}~\cite{di2017mining} such as histopathology, radiology, and microscopy—diagnostic tasks depend on subtle texture differences to identify tumors, tissue abnormalities, and cell boundaries, where explicit object edges are often absent. Similarly, in materials science~\cite{barba2018metallography, rusanovsky2022end, rusanovsky2023universal, cohen2024avoiding} , metallography and composite analysis require texture-based segmentation to detect grain structures, fractures, and surface irregularities, as these features lack clear semantic categories. In remote sensing, texture is essential for land cover classification and vegetation mapping, while in industrial inspection, defect detection relies on analyzing complex spatial statistics rather than discrete object boundaries. Despite the broad relevance of texture-based segmentation, prior methods have relied on ad-hoc models trained on often small, domain-specific datasets, limiting generalization. No foundation model has been established to provide a unified, transferable texture representation across domains. This work addresses this gap by introducing a protocol that adapts a pretrained foundation model for segmentation into a texture-aware model, enabling the use of learned texture representations across diverse segmentation tasks.

Existing segmentation models, including SAM~\cite{kirillov2023segment} and its successor SAM-2~\cite{ravi2024sam}, struggle in texture-driven applications where boundaries are defined by local surface properties rather than distinct object contours. This limitation highlights a fundamental research gap: can a generic foundation model for segmentation be developed that explicitly incorporates texture information rather than relying solely on shape-based priors?

We suspect that SAM-2 couples shape and texture cues. That is, it learns to expect that certain objects often have certain texture. Hence, our goal is to try and decouple the two.
To do that, we rely on a recently introduced compositional neural texture approach (CNT)~\cite{Tu:2024:CNT}, that can interpolate between a source and style (i.e., texture) images. This way, we can rely on ground truth masks of existing datasets (ADE20k in our case) to dress instances with arbitrary new texture, thus decoupling shape from texture.
Equipped with this new and augmented dataset we finetune SAM-2. We call the resulting model, {\em TextureSAM}.
We report the details of our method, and evaluate TextureSAM on a number of datasets.

% This work introduces TextureSAM, a fine-tuned foundation model that enhances segmentation performance in texture-dominant scenarios. By leveraging a texture augmentation strategy based on the method proposed in~\cite{Tu:2024:CNT}, a texture-altered training set is constructed where instance textures are systematically modified. Fine-tuning SAM on this dataset encourages it to rely on texture-defined regions rather than shape priors. This approach is the first to explicitly guide a foundation segmentation model to prioritize texture cues, addressing a major limitation of current methods.

\noindent\textbf{Contributions.} The key contributions of this work are as follows:
\begin{itemize}[left=0pt]
\item A texture-augmented ADE20K dataset is constructed to fine-tune SAM-2 as well as to provide benchmark segmentation performance in texture-aware scenarios.
\item We present a novel fine-tuning approach to shift SAM to focus on texture cues.
\item A fine-tuned segmentation model, TextureSAM, is introduced to enhance performance in texture-driven segmentation tasks. 
\item Extensive quantitative and qualitative evaluations demonstrate that TextureSAM significantly outperforms SAM-2 on both natural and synthetic texture-based segmentation datasets. 
\end{itemize}

\noindent\textbf{Research Questions.} This work investigates the following key research questions:
\begin{itemize}[left=0pt]
    \item \textbf{RQ1:} Is Segment Anything shape-biased? Can it comprehend textures?
    \item \textbf{RQ2:} Can a shape-biased foundation model be shifted towards texture-driven segmentation?
\end{itemize}

The remainder of this paper, is organized as follows: Section 2 reviews related work Section 3 details the fine-tuning approach and texture augmentation strategy. Section 4 presents experimental results and comparisons. Section 5 discusses limitations, applications, and future research directions.

\section{Prior Work}

\subsection{Texture Analysis}
Texture analysis is often associated with semantic segmentation and can be roughly divided into two parts: Semantic Segmentation and Instance Segmentation. In Semantic Segmentation, each pixel is assigned a label of the class (i.e., road, vehicle, sky, person, etc.) it belongs to. In Instance Segmentation, each pixel is assigned a label of the class and the instance it belongs to (i.e., two different people will be assigned different labels).

Recently, the two have been combined into a single, holistic approach, termed Panoptic Segmentation, where the goal is to assign each pixel its class label, and in case there are multiple instances of the object, assign the different objects a different label. Panoptic Segmentation is used for various applications such as Autonomous Driving and general image understanding.

Another closely related branch of texture analysis is that of Boundary Detection, where the goal is to detect meaningful boundaries/interfaces between textures, cells, crystallites, etc. This branch evolved from the problem of edge detection, where edge detection is concerned with detecting changes in intensity values, while boundary detection is concerned with detecting boundaries between regions (i.e., textures).

Traditional methods mostly focus on low-level cues. For example, Martin {\em et al.} \cite{Martin2004} proposed a probabilistic boundary (Pb) detection module that learns to detect boundaries in images using local brightness, color, and texture cues. Dollár and Zitnick~\cite{Dollar2014} use Structured Forests to map local image patches to local edge maps.

% A long line of work focused on extending the local approach to take global cues into consideration as well. 

A sequence of papers~\cite{DeepEdge,DeepContour,Kokkinos2015PushingTB} suggested using deep features to improve image boundary detection. Common to all is the use of deep features as the space in which to detect boundaries.
Exploiting multi-scale representations and dilated convolutions helped He {\em et al.}~\cite{BDCN} push the state-of-the-art in the field even further. Recently, Pu~{\em et al.} adapted Transformers for boundary detection~\cite{Pu_2022_CVPR}. Their approach is based on a two-level Transformer architecture, where the first level captures global scene information, and the second refinement level calculates the boundaries. Their network achieves state-of-the-art results on several datasets~\cite{jing2022recent}, including the BSD dataset~\cite{ArbelaezMFM11}. The latest advancement using Transformers for this task is the Segment Anything Model (SAM)~\cite{kirillov2023segment}, which presents a significant step forward while also having limitations in professional data, especially in textural and medical imaging~\cite{ji2023segment,huang2023segment}.

\subsection{SAM’s Limitations with Texture}

Segment Anything (SAM)~\cite{kirillov2023segment} introduced a new paradigm in segmentation research by enabling zero-shot segmentation across diverse datasets. SAM is trained on the SA-1B dataset, which contains one billion masks from eleven million images, using a Vision Transformer (ViT) backbone for feature extraction. SAM-2~\cite{ravi2024sam} extends this approach with a larger dataset and architectural refinements but remains fundamentally designed for semantic segmentation rather than texture-aware segmentation. ~\cite{huang2024segment} evaluate SAM for medical image segmentation, showing that it struggles with fine-grained structures and low-contrast regions, highlighting the need for domain adaptation. Prior work has tried to produce domain specific models. For instance~\cite{Ma2024Segment} introduces MedSAM, a model trained on a massive medical segmentation dataset. Their method attempts to address SAM-2's limitation with medical data, without directly addressing the issue of innate shape bias that is introduced by semantic-centric training data. This method is of course not applicable in domains that lack the massive annotated dataset required.

% designed to enable universal medical image segmentation, attempting to address SAM-2 attempting to bridge the gap between general image segmentation models and the specific requirements of medical imaging, but without directly addressing the issue of innate shape bias that limits pretrained foundation models from generalizing to other texture-driven domains. These domains often lack the immense datasets required for the creation of a domain-specific foundation model. 

The trade-off between shape and texture bias in deep learning has been extensively studied. Convolutional neural networks (CNNs) exhibit a strong texture bias due to their local receptive fields and weight-sharing properties~\cite{geirhos2018imagenet}. In contrast, vision transformers (ViTs), such as those used in SAM, prioritize global shape over local texture~\cite{naseer2021intriguing}. This is due to their patch-based tokenization and self-attention mechanisms, which emphasize long-range dependencies at the cost of fine-grained texture representations~\cite{dosovitskiy2020image}. Experiments using Stylized ImageNet~\cite{naseer2021intriguing} confirm that ViTs retain object recognition even under extreme texture alterations, further reinforcing their preference for shape-based segmentation. Additionally, real-world evaluations of SAM-2~\cite{ji2024segment} reveal that its performance degrades in texture-sensitive scenarios, underscoring the need for improved texture awareness in segmentation foundation models. 
% Zhang et al.~\cite{zhang2023understanding} analyzed the Segment Anything Model (SAM) and concluded that it is biased toward texture rather than shape, contrasting with prior findings that transformer-based models exhibit a shape bias and diverging from our own results, which indicate that SAM primarily relies on shape cues for segmentation. One explanation for this discrepancy, is that their study used a limited set of image types, including silhouettes and edge images, which strip away color, shading, and fine structural details present in real-world images.

Recent work has explored methods for modulating shape versus texture bias in vision models. Gavrikov et al.~\cite{gavrikov2024vision} demonstrated that shape bias in vision-language models can be controlled via language prompts, but to our knowledge, no prior work has explicitly addressed increasing texture bias in segmentation foundation models. While texture transfer techniques such as~\cite{Tu:2024:CNT} enable controlled modification of textures, they have not been used to systematically fine-tune a foundation model for texture-aware segmentation. This work introduces a texture-focused adaptation of SAM, directly addressing its shape-prior limitation.

 % Moreover, studies have shown that SAM struggles in applications requiring precise texture differentiation~\cite{ji2024segment}, reinforcing the need for explicit texture modeling in foundation models.

\section{TextureSAM}

\subsection{Overview}

TextureSAM is a texture-aware variant of the Segment Anything Model (SAM) that is created by finetuning it on texture-augmented datasets. While SAM-2 excels at general segmentation, its reliance on high-level semantic cues limits its performance in scenarios where texture is the primary distinguishing feature. To address this, we finetune SAM-2 on ADE20K, augmenting images with a state-of-the-art texture replacement method. Textures are incrementally introduced within object regions defined by ground-truth masks, using samples from the Describable Textures Dataset (DTD) ~\cite{cimpoi2014describing}, which is a dataset designed for studying texture recognition in real-world, unconstrained environments  containing 5,640 texture images. This augmentation encourages SAM-2 to leverage texture cues for segmentation.

We train two versions of TextureSAM: one with mild texture augmentation ($\eta \leq 0.3$), which preserves most semantic structure, and another with strong texture augmentation ($\eta \leq 1.0$), where objects are fully replaced by textures, eliminating all semantic information. 

We evaluate TextureSAM against the original SAM-2 on two texture-focused datasets: RWTD~\cite{Khan2018LearnedShapeTailored}, a natural image dataset with texture-based segmentation ground truth, and a synthetic dataset, STMD~\cite{Mubashar2022EdgeDetection}, composed solely of texture transitions. Performance is measured using mean Intersection over Union (mIoU) and Adjusted Rand Index (ARI) to assess segmentation quality under texture-based conditions.

\newcommand{\imageSize}{0.14\linewidth}

\begin{figure*}[ht]
    \centering
    \begin{tabular}{cccccc}    \includegraphics[width=\imageSize]{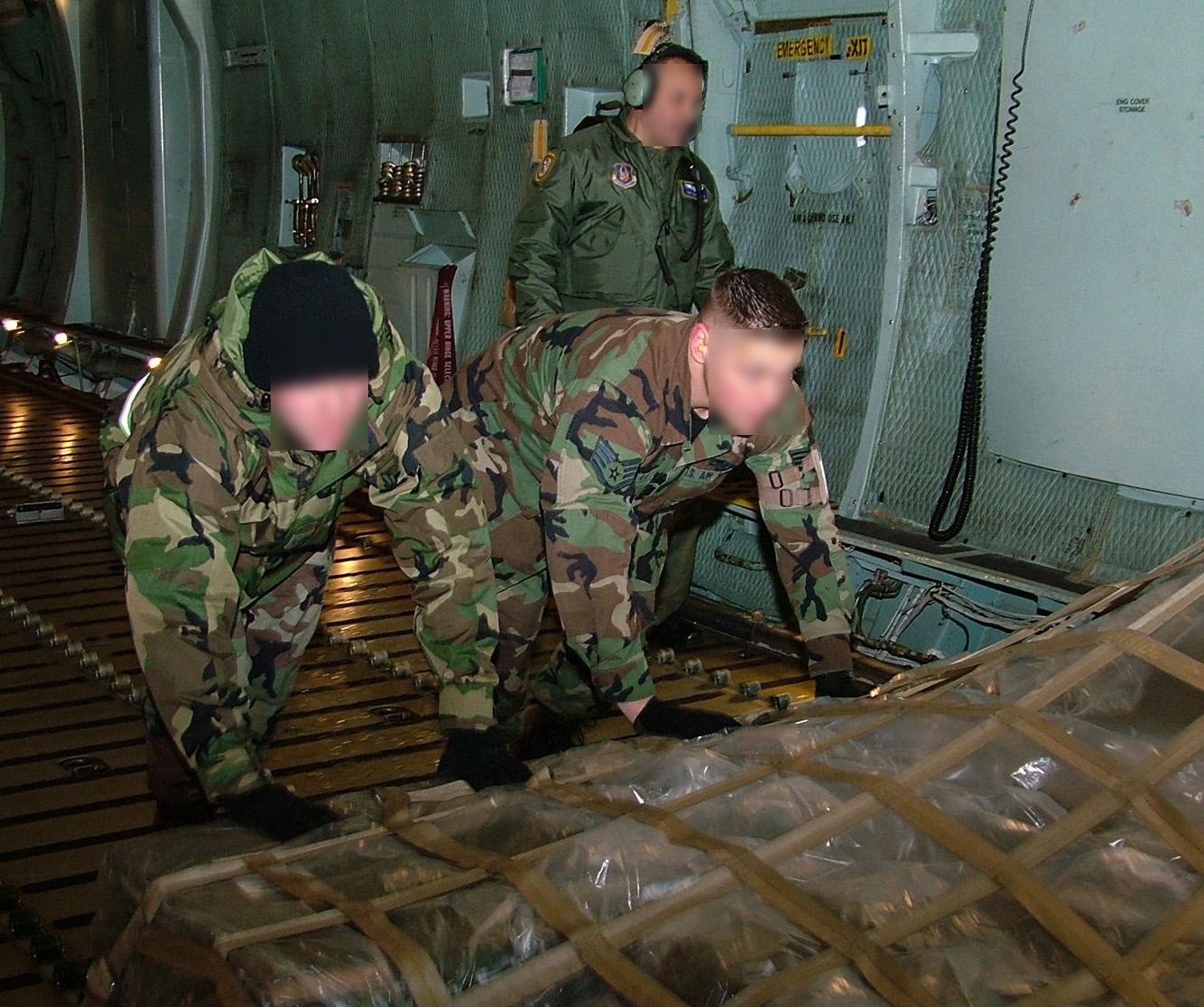} &
    \includegraphics[width=\imageSize]{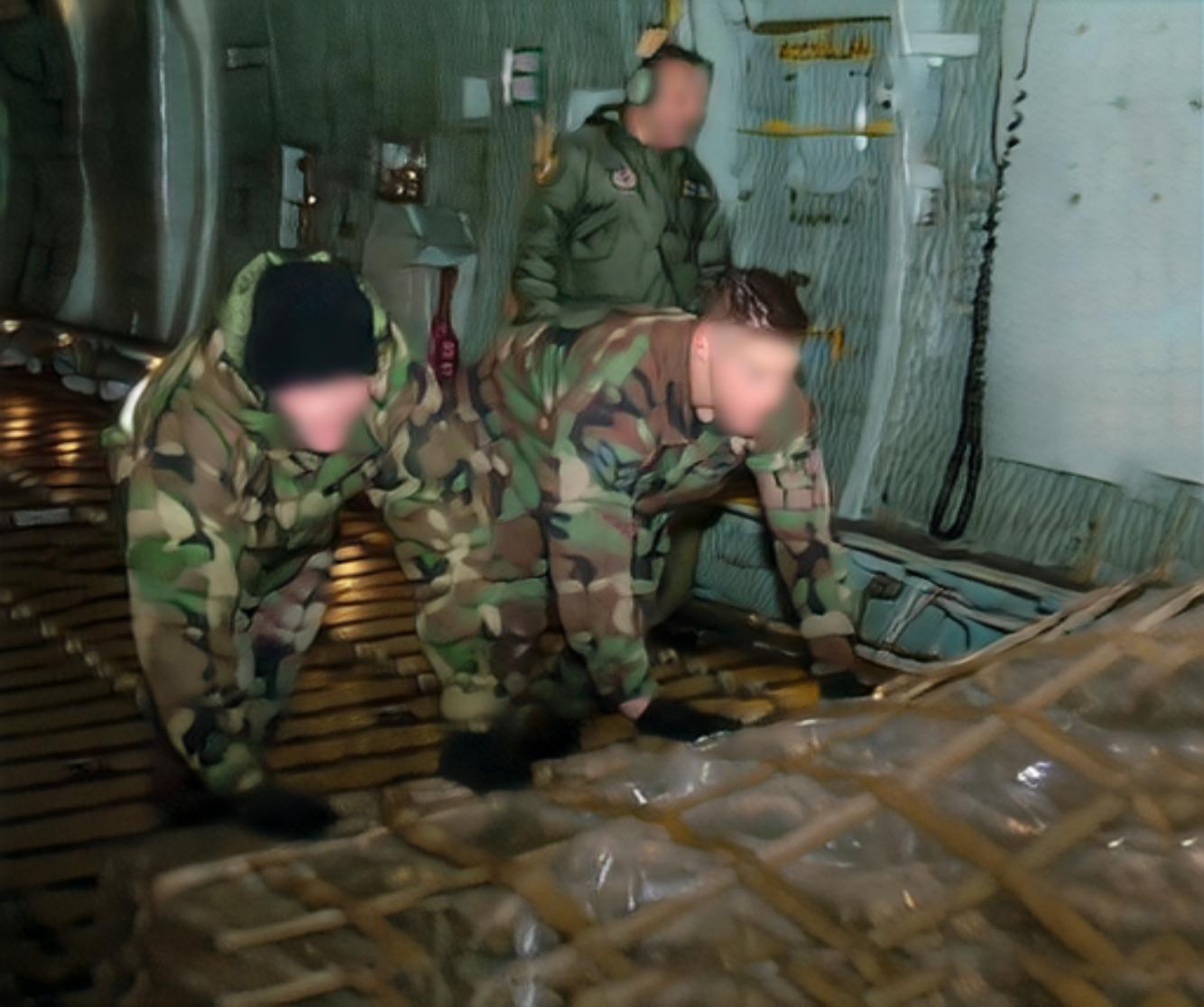} &
    \includegraphics[width=\imageSize]{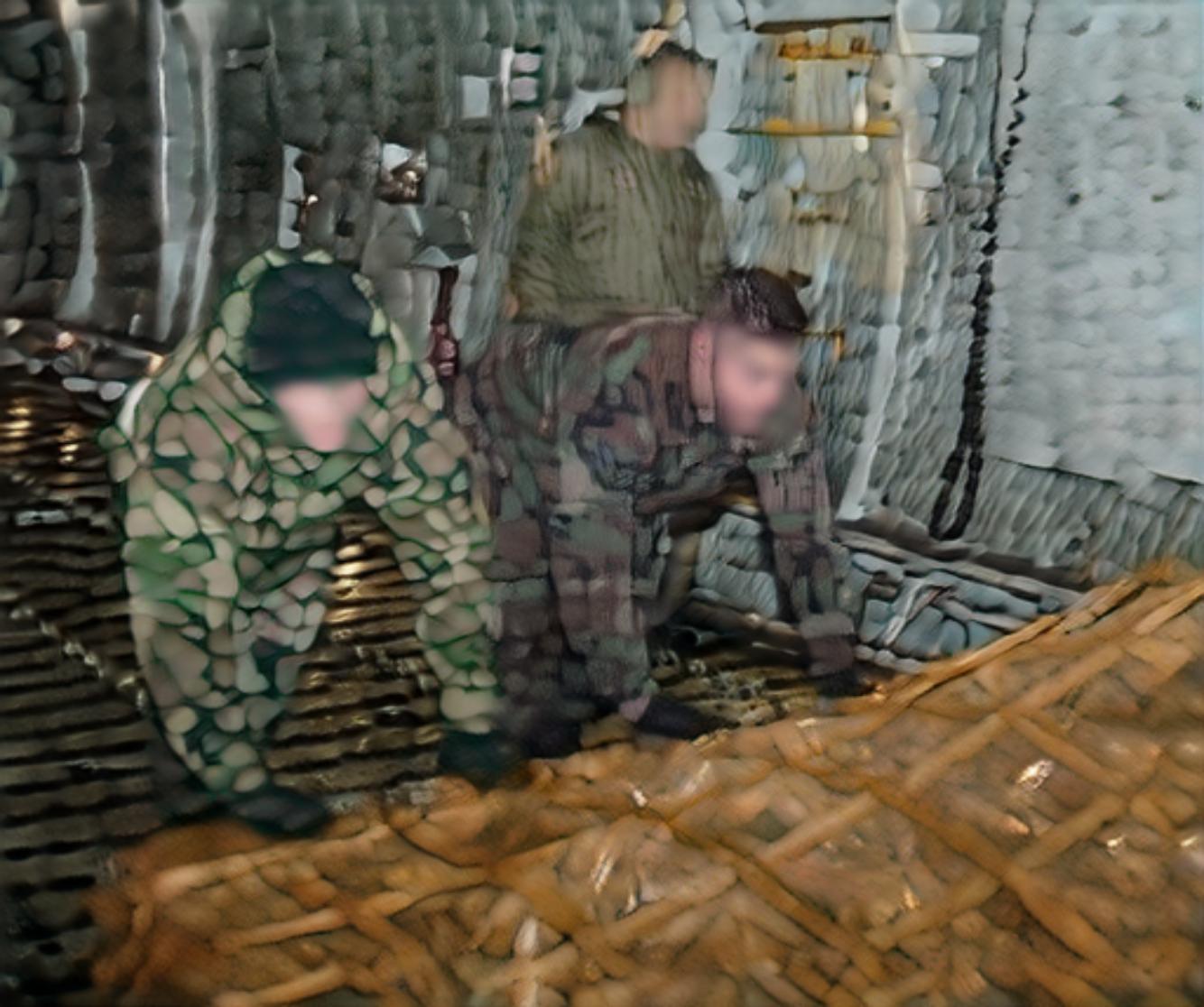} &
    \includegraphics[width=\imageSize]{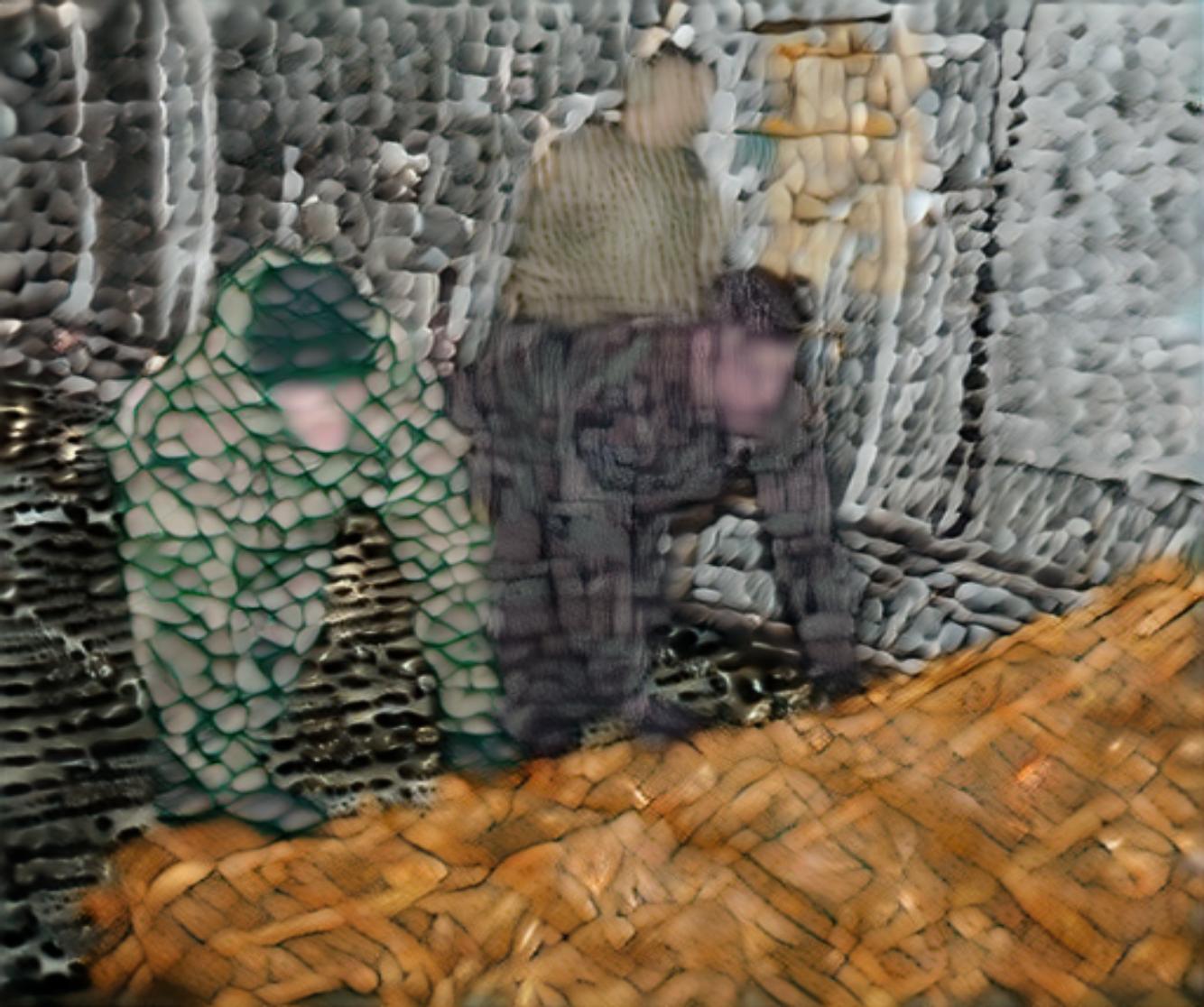} &
    \includegraphics[width=\imageSize]{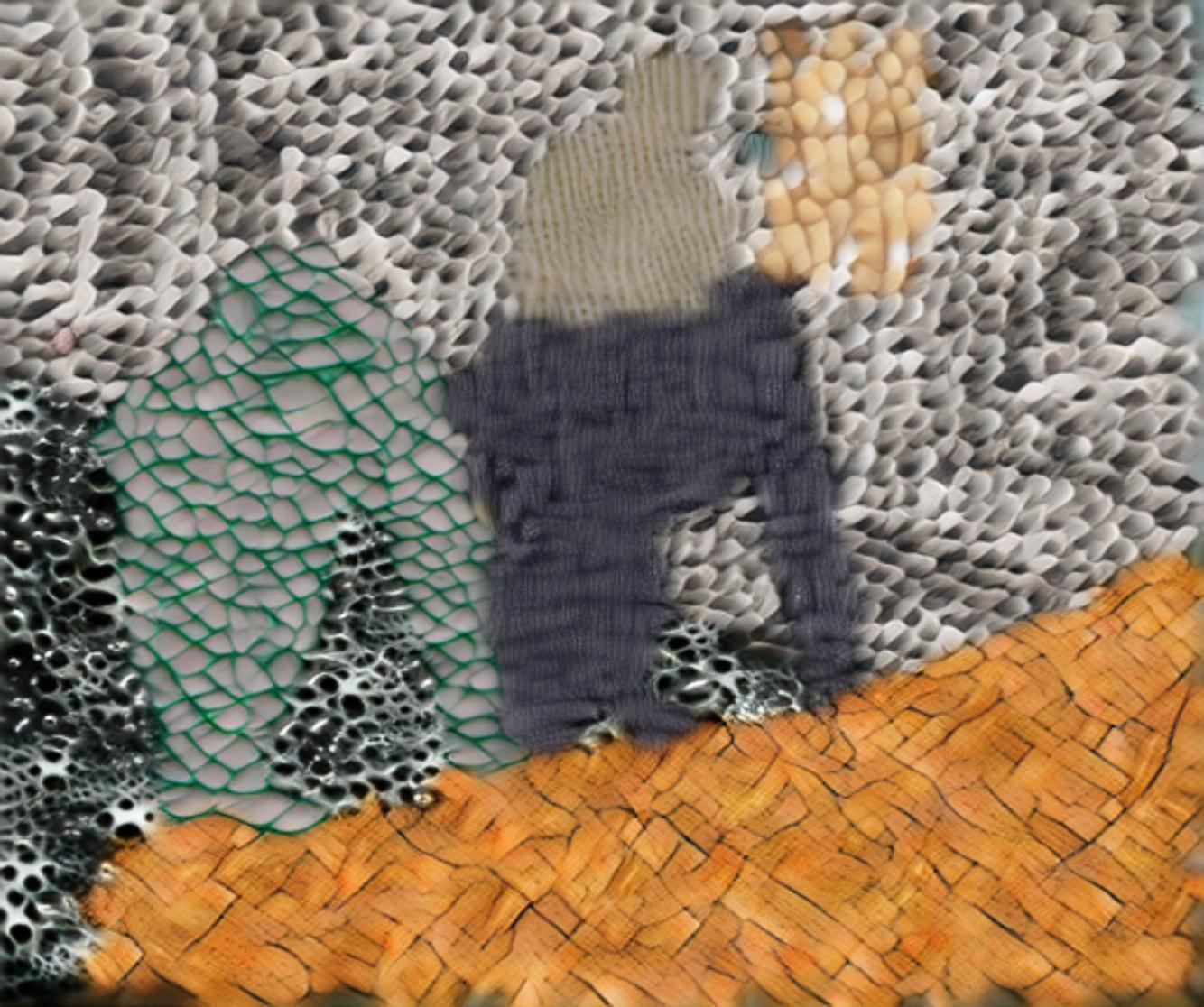} &
    \includegraphics[width=\imageSize]{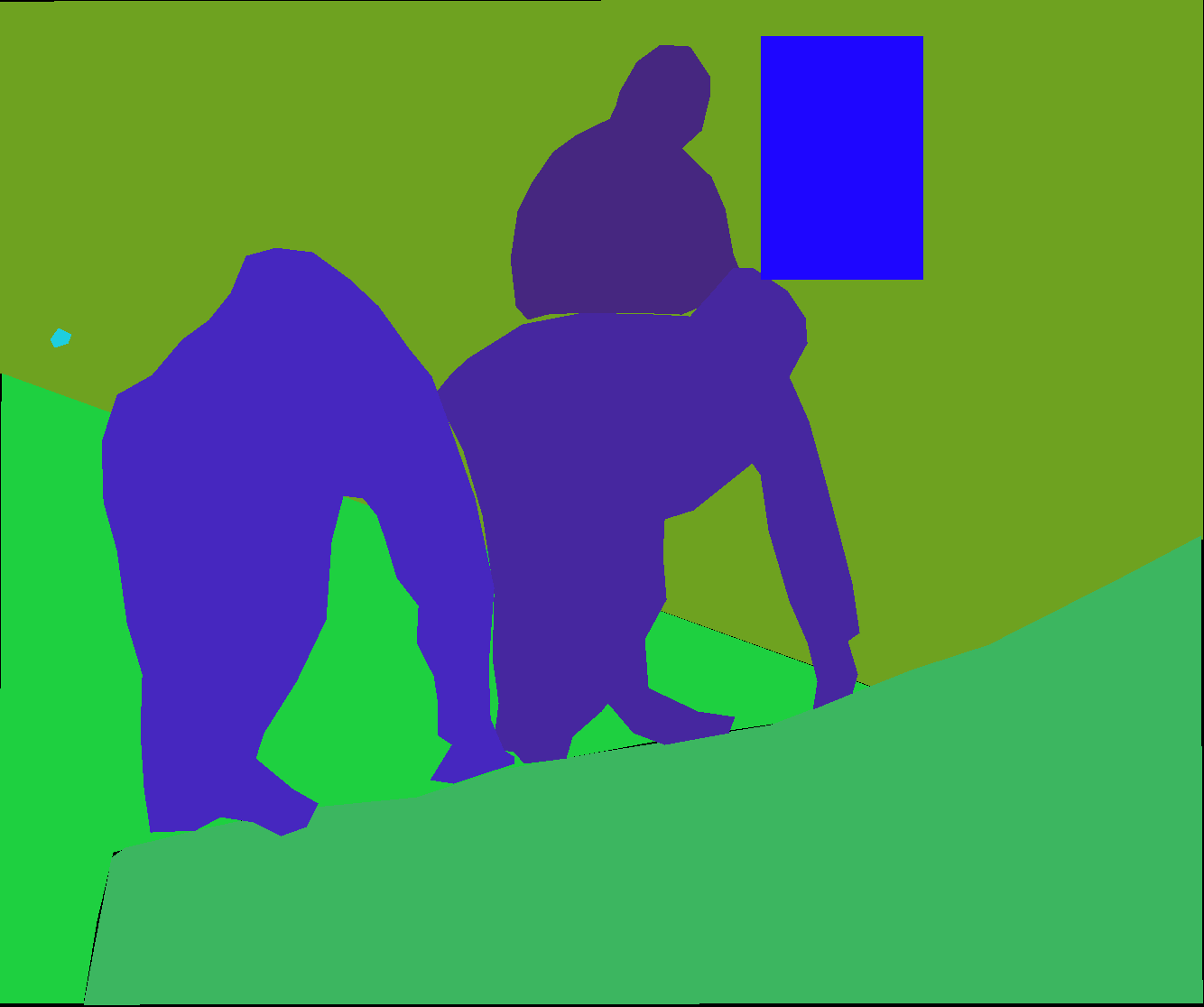} \\
   
    %%%%%%%%%%%%%%%%%
    \includegraphics[width=\imageSize]{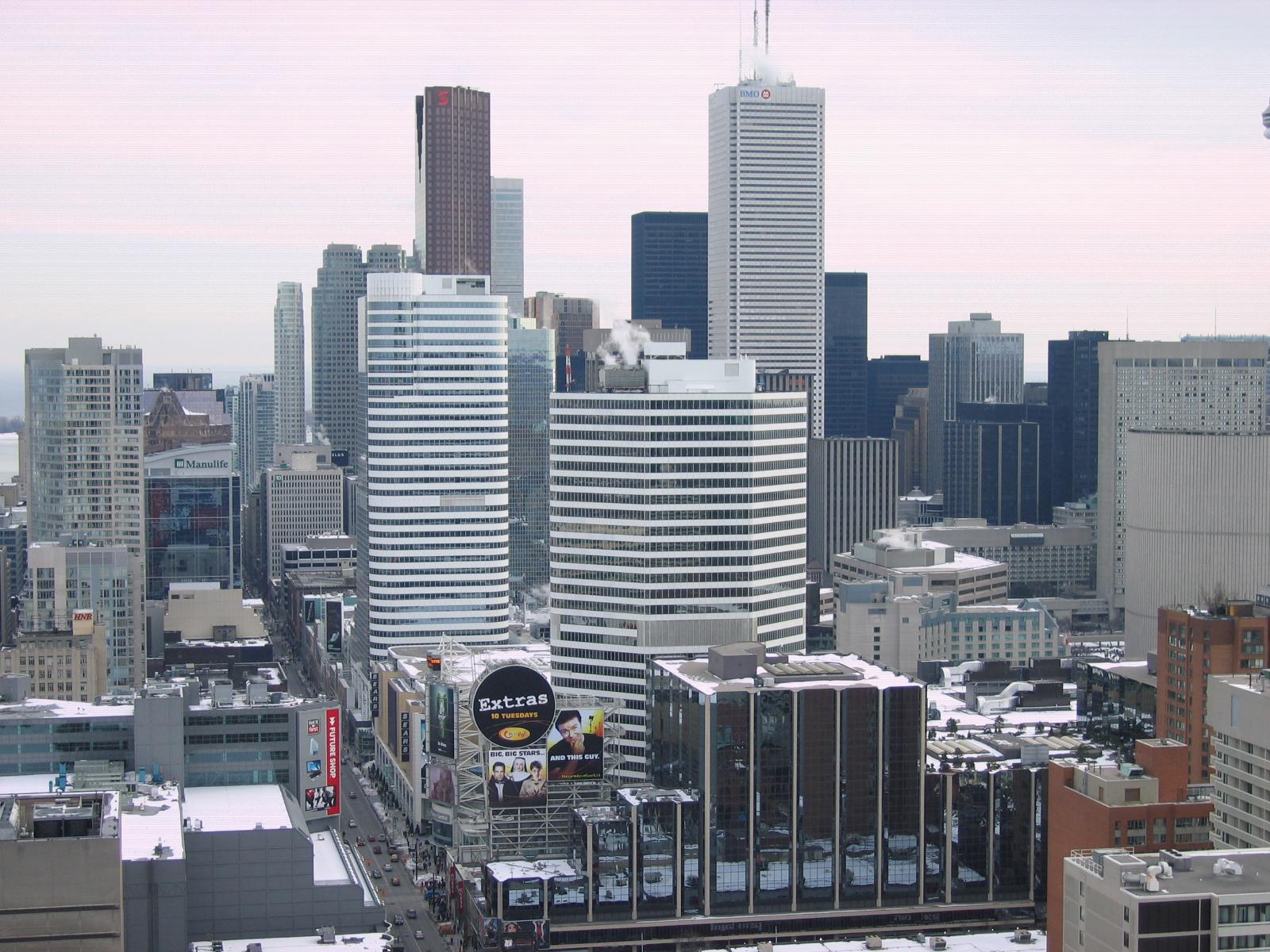} &
    \includegraphics[width=\imageSize]{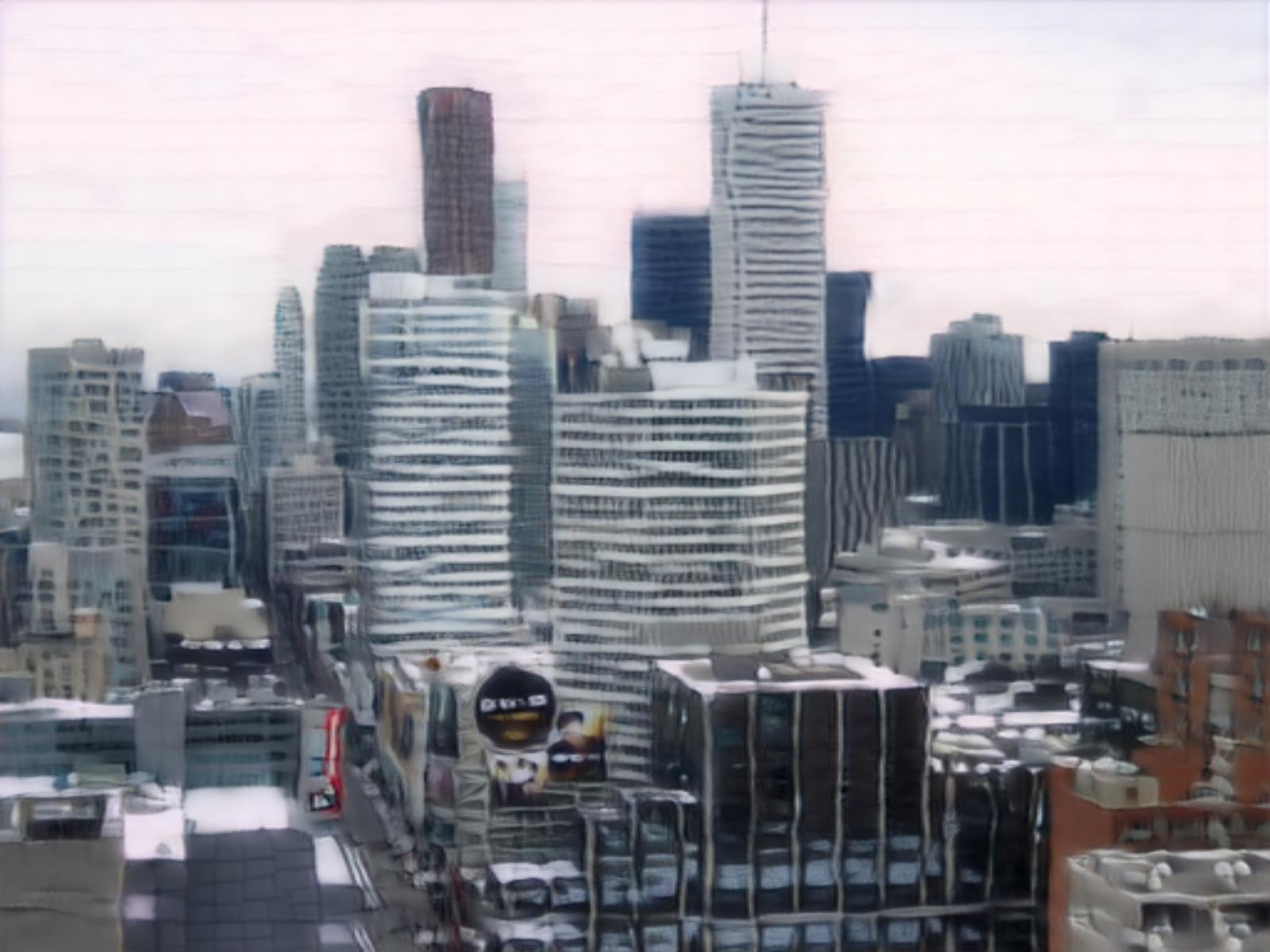} &
    \includegraphics[width=\imageSize]{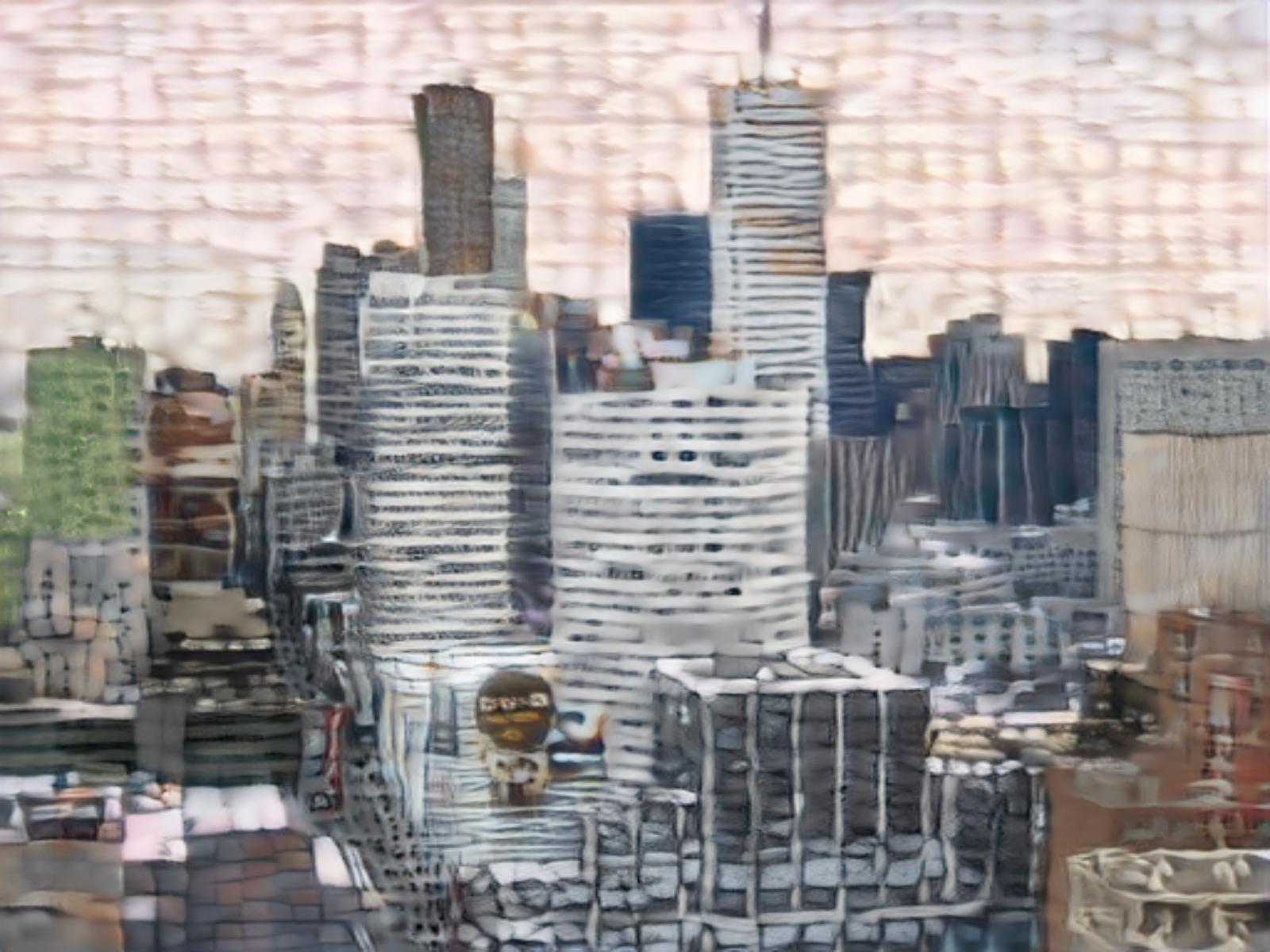} &
    \includegraphics[width=\imageSize]{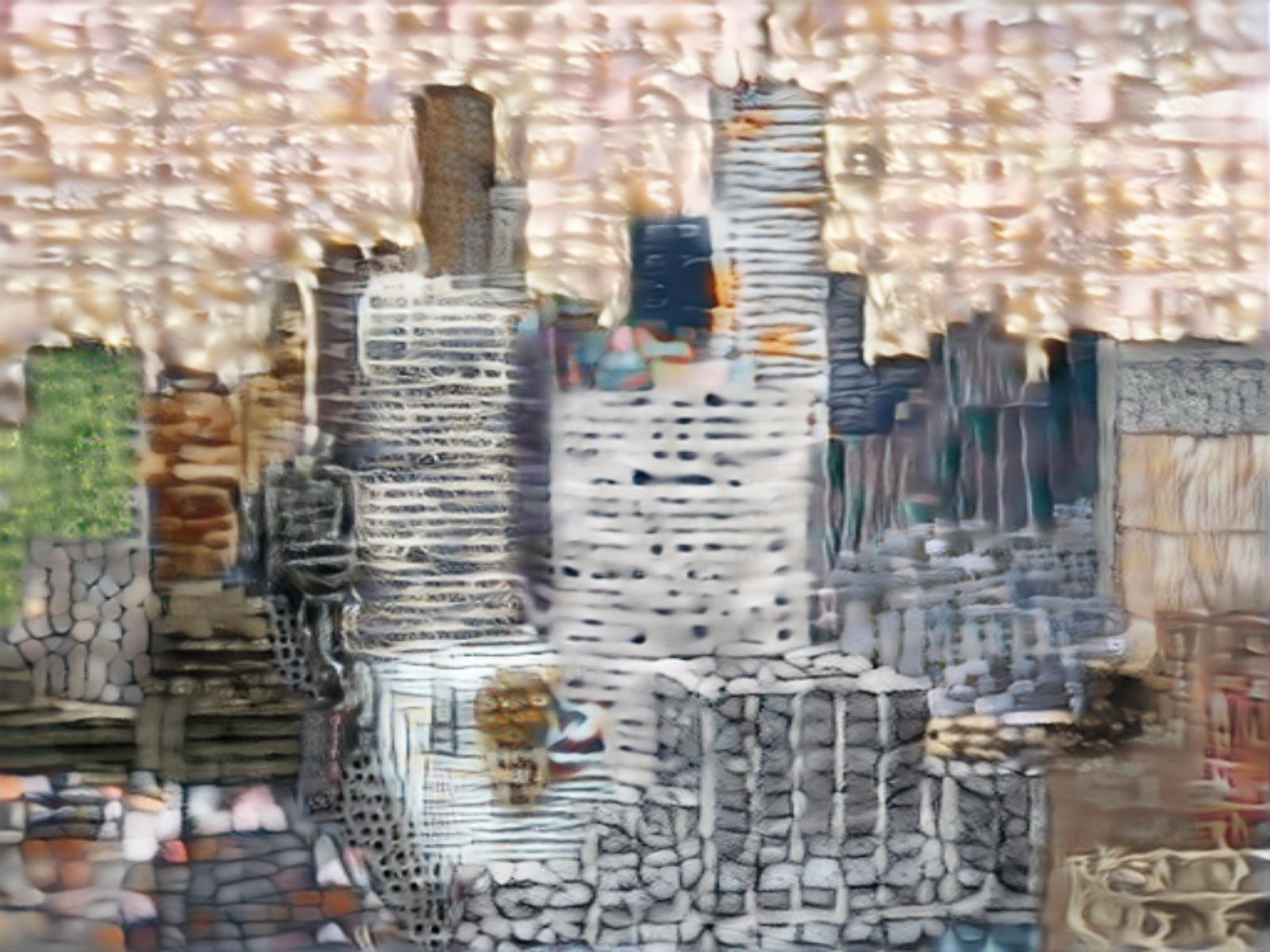} &
    \includegraphics[width=\imageSize]{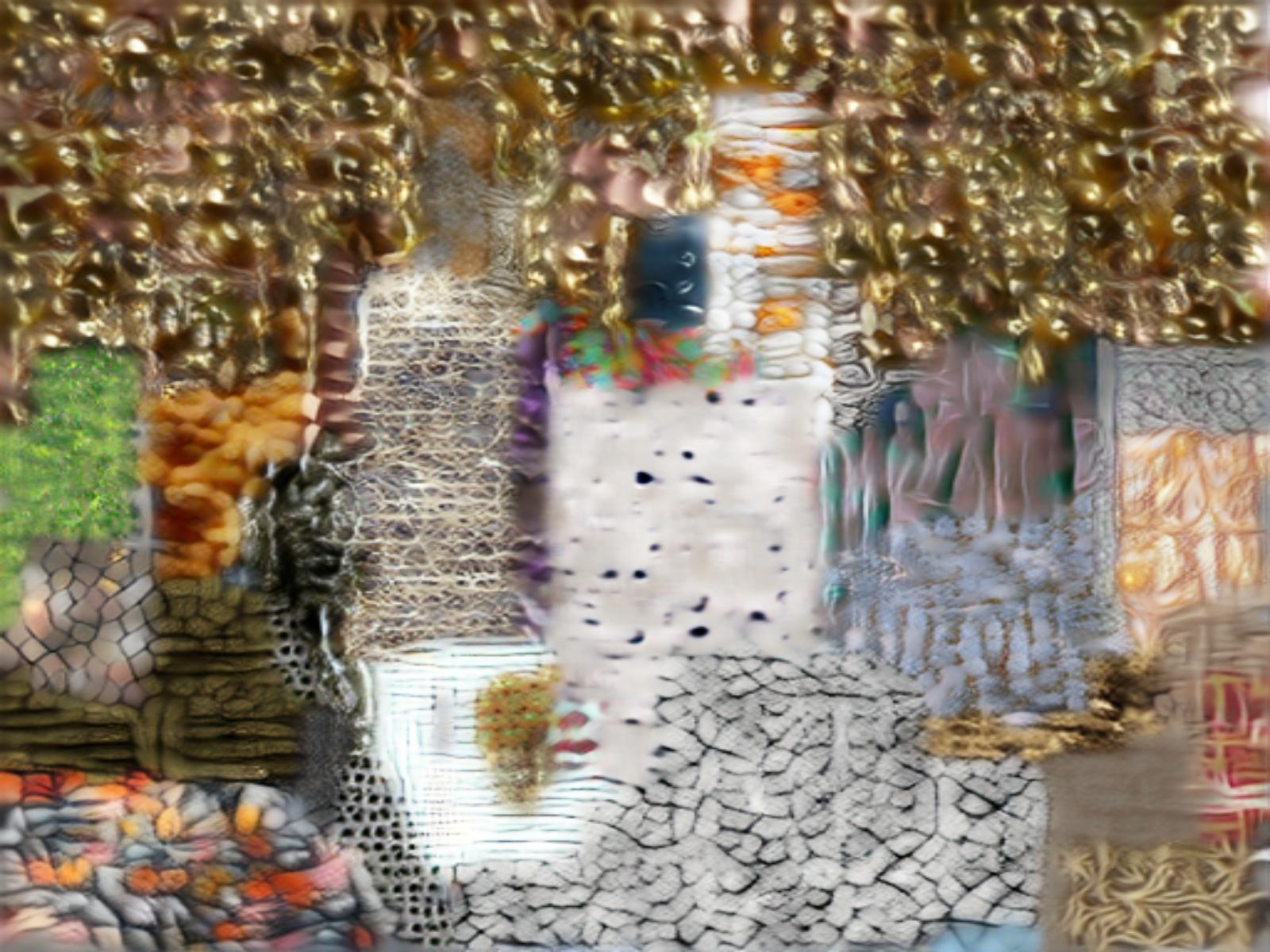} &
    \includegraphics[width=\imageSize]{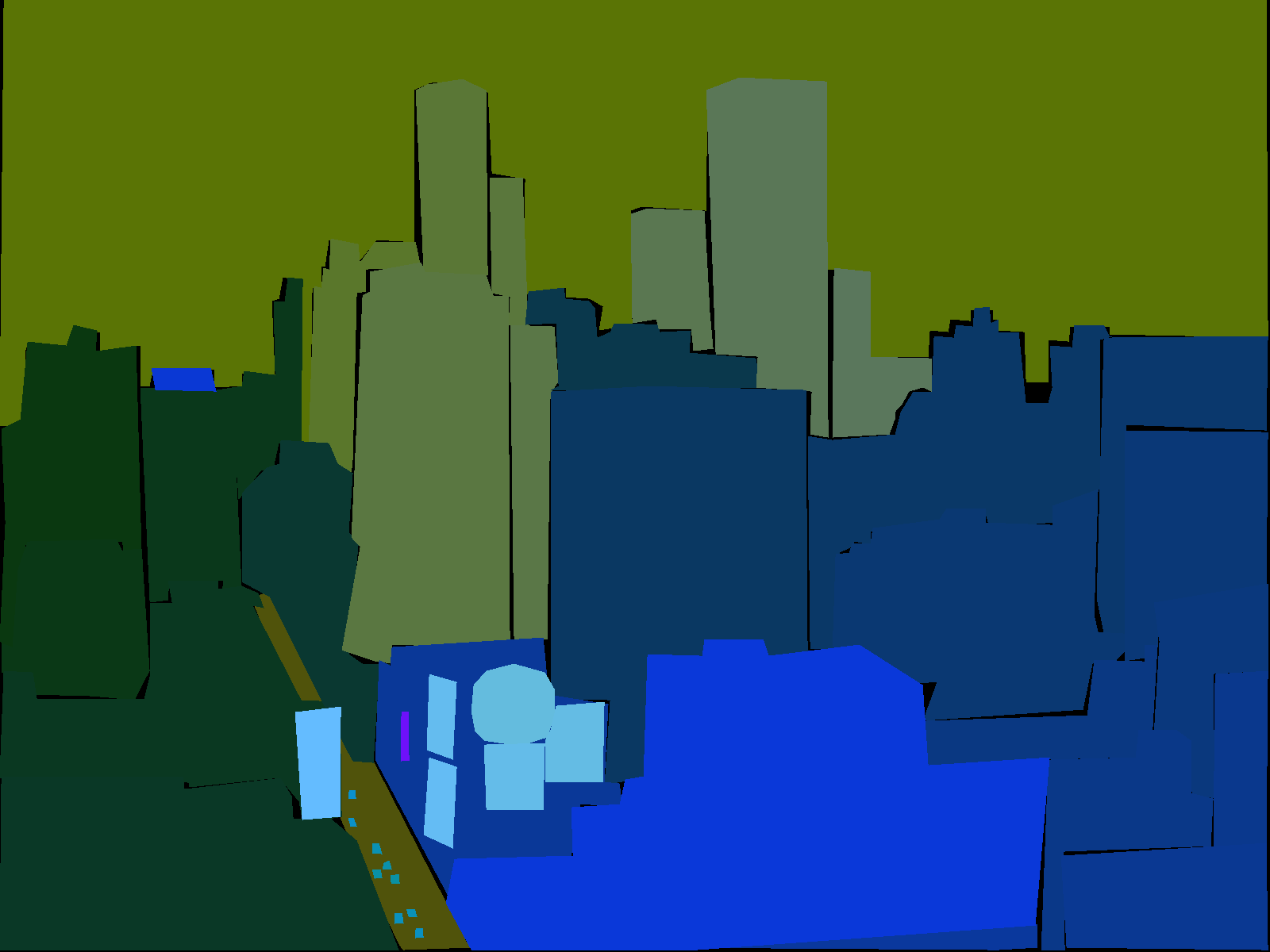} \\

    Original & $\eta=0.0$ & $\eta=0.4$ & $\eta=0.6$ & $\eta=1.0$ & Instance mask \\

    \end{tabular}
    
    \caption{Samples of Textured-ADE20K dataset. Incremental changes in $\eta$ produce gradual change in texture shift for the resulting image. For low $\eta$ values most of the semantic information in the image is retained. For $\eta=1$ the instances are completely shifted towards the target textures.}
    \label{fig:dataset_samples}
\end{figure*}

\subsection{Datasets}

The ADE20K dataset ~\cite{zhou2017scene} is a large-scale scene parsing dataset containing over 20,000 images with 150 semantic categories, including objects, parts, and materials. It provides densely annotated scene layouts, making it a benchmark for semantic segmentation, scene understanding, and contextual reasoning. The dataset covers a diverse range of indoor and outdoor environments, making it widely used in deep learning-based segmentation research.
The training part of this dataset was used to train TextureSAM, and the validation part of the ADE20K dataset, comprising 2000 images, is used to assess our models ability to perform general semantic segmentation after fine-tuning. 

To assess the effectiveness of TextureSAM, we evaluate its texture segmentation performance on two texture-centered datasets where boundaries are defined by texture changes rather than object semantics.

We use the Real World Textures Dataset (RWTD), a natural image dataset containing 256 annotated images where the ground truth marks texture boundaries rather than object edges. This dataset originates from ~\cite{Khan2018LearnedShapeTailored} and provides a challenging benchmark for evaluating segmentation models that rely on texture cues.

In addition, we evaluate on the Synthetic Textured Masks Dataset (STMD), introduced in~\cite{Mubashar2022EdgeDetection}. Unlike RWTD, STMD contains no explicit objects, consisting only of synthetic images with clear texture transitions. This dataset isolates texture-based segmentation performance by eliminating shape and semantic information, making it a strong test case for assessing TextureSAM’s ability to differentiate regions based purely on texture.

% \subsection{Dataset Preparation}

% \paragraph{ADE20K for Finetuning}
\paragraph{Dataset Preparation}
To finetune SAM-2 while preserving its original segmentation capabilities, we use the ADE20K training set which is a subset of SA-1B, the original data used in SAM-2's training. This approach mitigates catastrophic forgetting \cite{french1994catastrophic}, ensuring that TextureSAM retains general segmentation ability while adapting to texture-based cues. ADE20K presents a particularly challenging benchmark, with the original SAM-2 achieving only $0.46$ mIoU on its validation set, indicating ample room for improvement. The following section details the augmentation strategy applied to ADE20K dataset in order to introduce texture awareness in SAM-2.
% We retain the standard ADE20K validation set to monitor segmentation performance at different levels of texture augmentation ($\eta \leq 0.3$ and $\eta \leq 1.0$). This validation set allows us to assess whether the model progressively shifts towards texture-dependent segmentation while maintaining its general segmentation ability (see Supplemental Materials for detailed validation results).

% \paragraph{Texture Augmentation}

% To introduce texture awareness into SAM, we augment the ADE20K training images using a state-of-the-art texture replacement method. Textures are sampled from the DT dataset and applied incrementally within semantic regions defined by ground-truth masks. The augmentation is controlled by $\eta$, which determines the degree of texture replacement. At $\eta \leq 0.3$, semantic structures remain largely intact, while at $\eta \leq 1.0$, objects are fully replaced by textures, eliminating all semantic information. A full description of the augmentation method, including implementation details and examples, is provided in Section 4: Augmented ADE20K Dataset.

\subsection{Textured-ADE20K dataset}

We employ the texture transfer technique in ~\cite{Tu:2024:CNT} to create the Textured-ADE20K dataset. We transfer textures $\textureimage$ from describable texture  dataset (DTD)~\cite{Cimpoi:2014:DTW} to semantic content images $\contentimage$ from ADE20K ~\cite{Zhou:2017:ADE}.
There are 5640 texture images organized into 47 categories of textures in the DTD dataset. ADE20K dataset contains 27574 images (25574 for training and 2000 for validation), where each image comes with associated ground-truth instance segmentation masks $\maskset$.
To create the textured-ADE20K dataset for segmentation, the instances in an image are transferred with different textures. Fig.~\ref{fig:dataset_samples} shows samples from the textured-ADE20K dataset.

%%%%%%%%%%%%%%%%%%

\begin{figure}[t]
    \centering
    \captionsetup[subfloat]{labelformat=empty}
    \subfloat[]{%
    \includegraphics[width=1\linewidth]{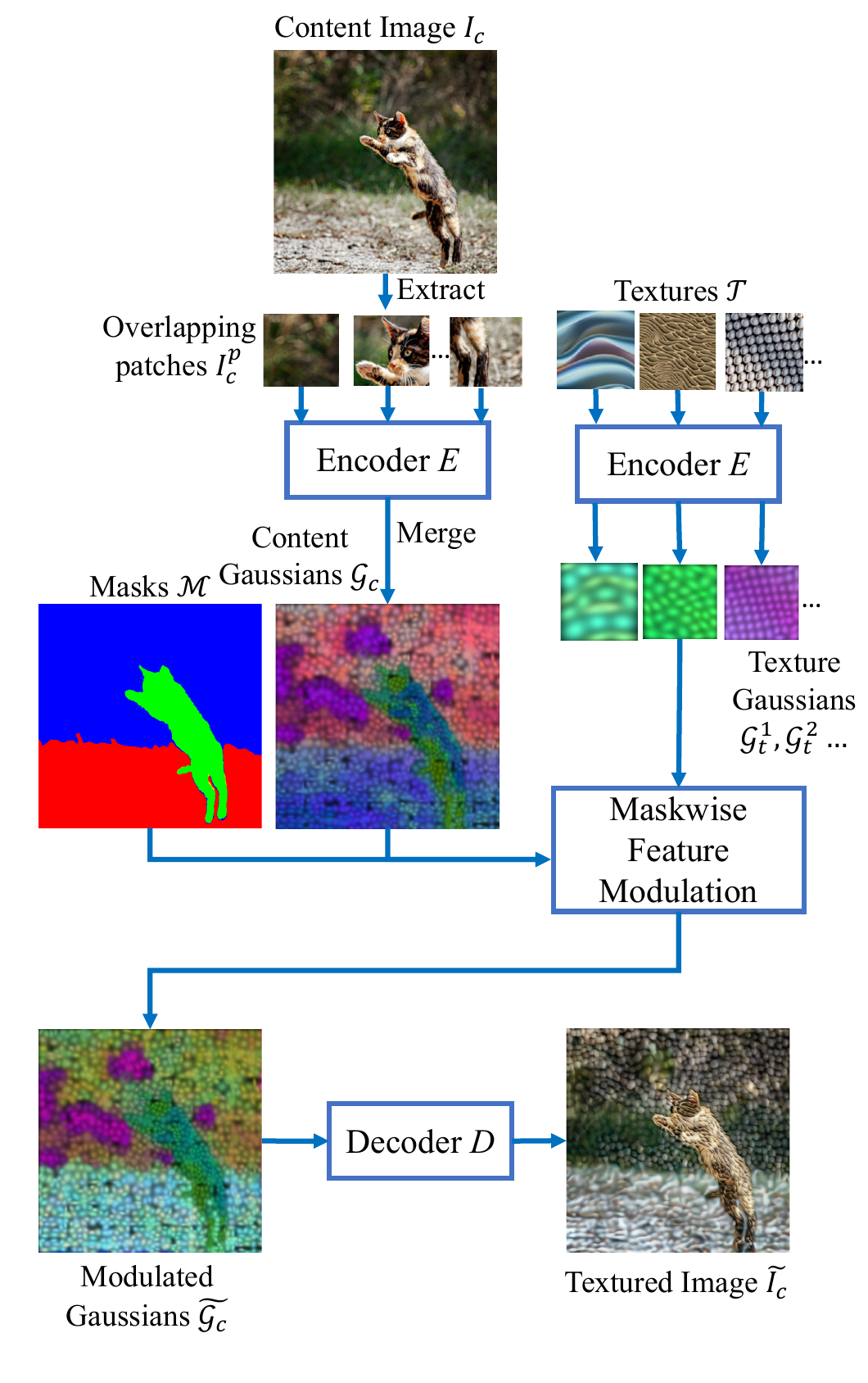}%
    }%
    \caption{Illustration of generating textured image for dataset augmentation using \cite{Tu:2024:CNT}.}
    \label{fig:texture_transfer_method}
\end{figure}

Figure~\ref{fig:texture_transfer_method} gives an overview of the method. To texturize an image $\contentimage$, for each instance mask $\mask$, we randomly select a texture $\textureimage$ from a unused category in the DTD dataset (if all categories are used, randomly select a texture from the entire dataset).
Before further processing, we scale the content image $\contentimage$ into $8\imagesize\times8\imagesize$ pixels and the texture image $\textureimage$ into $\imagesize\times\imagesize$ pixels.
Since the model focuses solely on texture, we extract overlapping patches $\contentimagepatch$ of size $\imagesize\times\imagesize$ with spacing of $3/4\imagesize$ pixels from $\contentimage$, $\patch \in \patchset=\setof{0,1,2,...,\numpatches}$ -- treating every patch as an independent texture sample.
$\contentimagepatch$ is then encoded into a composition of Gaussians $\gaussiansetcontentpatch$, and these representations are merged in their original spatial order to produce Gaussians $\gaussiansetcontent=\merge(\bigcup_{\patch \in \patchset}\gaussiansetcontentpatch)$ reconstructed as the original image. $\merge$ merges overlapping patches of Gaussians while ensuring smooth transition (see \cite{Tu:2024:CNT} for details).
Each texture $\textureimage$ is also encoded as Gaussians $\gaussiansettexture=\setof{\gaussiantexture}$ by directly forwarding them through the encoder $\encoder$.
To alter texture within a binary instance mask $\mask \in \maskset$, we select Gaussians centered within the mask $\contentmasked{\gaussianset}=\setof{\contentmasked{\gaussian}}=\setof{\gaussiancontent | \gaussiancontent \in \gaussiansetcontent, \mask(\gaussiancenter_\contentsymbol)=1}$ and modulate their appearance features $\contentmasked{\appearancefeature} \in \contentmasked{\gaussian}$. For each $\contentmasked{\appearancefeature}$, we randomly pick a feature $\appearancefeature_\texturesymbol \in \gaussiantexture$ from $\gaussiansettexture$, the modified feature $\contentmasked{\modified{\appearancefeature}}$ is computed as
\begin{align}
\contentmasked{\modified{\appearancefeature}}&=\interpcoefficient \contentmasked{\appearancefeature} + (1-\interpcoefficient) \appearancefeature_\texturesymbol.
\end{align}

The interpolation coefficient $\interpcoefficient$ determines the tradeoff between the input image and the texture image.
When $\interpcoefficient=0$, $\modified{\contentimage}$ is the \emph{texturization} of $\contentimage$: $\modified{\contentimage}$ closely matches $\contentimage$ but is not the exact reconstruction (Fig.~\ref{fig:dataset_samples}, rows: 1-2, columns: 1-2).
This is because we encode $\contentimage$ as Gaussians using $\encoder$ patch-by-patch and reconstruct them using $\decoder$, which (1) projects each image patch onto the learned texture space (2) merges overlapping Gaussian patches and (3) reconstruct Gaussians as $\modified{\contentimage}$.
Intuitively, this process 1) finds closest texture embeddings for image patches 2) spatially merge texture embeddings to ensure smooth transition and 3) reconstruct images from blended textures.

% The interpolation coefficient $\interpcoefficient$ determines the tradeoff between the input image and the texture image.
% When $\interpcoefficient=0$, the texturized output image closely matches the input image, but is not an exact reconstruction (Fig.~\ref{fig:dataset_samples}, rows: 1-2, columns: 1-2).
% This is because we encode the texturedized image as Gaussians using an encoder patch-by-patch and reconstruct them using the decoder, which 1) projects each image patch onto the learned texture space 2) merges overlapping Gaussian patches and 3) reconstruct Gaussians as the texturized output image.
% Intuitively, this process 1) finds closest texture embeddings for image patches 2) spatially merge texture embeddings to ensure smooth transition and 3) reconstruct images from blended textures.

% We use an autoencoder to represent a texture as a Gaussian Mixture Model ~\cite{Tu:2024:CNT}. Figure~\ref{fig:texture_transfer_method} gives an overview of the method. Additional details are in the supplemental.

% We use the method of Tu {\em et al.}~\cite{Tu:2024:CNT} to texturize an image. This is done by embeding both the texture and the image into latent space, interpolate between them and decoding it back. Figure~\ref{fig:texture_transfer_method} gives an overview of the method. Additional details are in the supplemental.

% This is the description of the texture interpolation method
% \input{CNT}

\subsection{Model Finetuning}

The training protocol for TextureSAM follows the default finetuning configuration from the SAM-2 repository \footnote{\url{https://github.com/facebookresearch/sam2}}, ensuring consistency with prior work. Hyperparameters remain unchanged and are provided in the Supplemental Materials for reference. Due to resource constraints, we finetune the sam2\_hiera\_small variant of SAM-2 using a single A100 GPU. Despite the reduced model size, this configuration allows us to efficiently evaluate the impact of texture augmentation while maintaining alignment with the original training setup. We train for 19 epochs on $\eta \leq 0.3$ and 25 epochs on $\eta \leq 1.0$, balancing training time with performance improvements.

\subsection{Evaluation Protocol}

\paragraph{Metrics}

We measure segmentation performance using two complementary metrics:
\begin{itemize}[left=0pt]
    \item \textbf{Mean Intersection over Union (mIoU):} Evaluates the overlap between predicted and ground truth regions.
    \item \textbf{Adjusted Rand Index (ARI):} Measures clustering consistency, particularly useful for assessing segmentation quality in texture-based datasets, as it penalizes fragmentation of textured regions to individual perceptual elements.
\end{itemize}

We compare TextureSAM ($\eta \leq 0.3$ and $\eta \leq 1.0$) against the original SAM-2, analyzing how well the model adapts to texture cues while maintaining general segmentation capability.

\paragraph{Inference and Evaluation Procedure.}

To ensure meaningful segmentation results, we modify the default inference parameters of SAM-2 for TextureSAM inference. Specifically, we modify the points\_per\_side parameter from 32 to 64, and the stability\_score\_thresh from 0.95 to 0.2. 
% \ref{fig:mod_parameters} qualitatively shows the affect the stability\_score\_theresh modification has on both the original SAM and TextureSAM. \ref{fig:mod_parameters} shows the same change but with points\_per\_side=64'. 
Using the default inference parameters for TextureSAM resulted in no predicted masks for most images, making direct evaluation unreliable. By adjusting the working point of the model inference we allow for a more meaningful comparison. Given that the modified inference parameters lead to a more dense segmentation which may increase textured area fragmentation, we ensure a fair evaluation by obtaining results for the original SAM-2 model with both the original and modified inference parameters. Throughout this paper, we refer to the original model with default parameters as \textbf{SAM-2}, and to the model evaluated with our adjusted inference parameters as \textbf{SAM-2*}.

\textbf{Predicted Mask Aggregation.} For each ground truth (GT) mask, we first identify overlapping predicted masks in the model’s output and unify them before calculating IoU. This provides an evaluation of the model's overall segmentation ablity, disregarding fragmentation. We report results both with mask aggregation and without.

% The evaluation is performed only on relevant predicted masks, ensuring that segmentation quality is assessed within corresponding texture-defined regions. mIoU of the relevant prediction masks is computed  for each GT region. We then report the average mIoU scores across all images in the dataset. Similarly, ARI is calculated based on the agreement between GT and predicted segmentation clusters. This approach ensures that evaluation reflects how well the model captures texture-based boundaries rather than partial texture elements.

% \paragraph{Considerations and Limitations}

% To ensure a fair comparison, for the original SAM we report both results obtained with the default inference parameters and the modified inference parameters used for TextureSAM.

% We evaluate generalization on RWTD, as there is no overlap between ADE20K (used for training) and RWTD. The ground truth (GT) in RWTD marks only a single texture boundary per image, even though about 5\% of images contain three or more texture changes. This means that some texture transitions are not explicitly accounted for in evaluation.

% The Synthetic Textured Masks Dataset (STMD) provides a more controlled benchmark for testing multiple texture boundaries per image. Unlike RWTD, STMD contains no semantic objects, ensuring that segmentation performance is based purely on texture differences. This allows for an exact evaluation of how well the model captures texture-based segmentation without interference from object structure.

The next section provides a detailed description of the augmented ADE20K dataset, including the texture augmentation method, dataset statistics, and example transformations.

\section{Results}

We evaluate our method using two challenging texture-oriented segmentation datasets: the RWTD natural images dataset, and a STMD dataset that consists of synthetic images with multiple texture changes and no foreground objects. Each dataset presents distinct challenges for texture-aware segmentation, enabling a comprehensive assessment of our approach. 

We compare our texture-aware model, TextureSAM, against the original Segment Anything Model (SAM2). To allow for a fair comparison, for the original SAM2 model we perform the evaluation with the default inference parameters, as well as the modified inference parameters used for TextureSAM. Evaluation is performed using two primary metrics: mean Intersection over Union (mIoU) and Adjusted Rand Index (ARI). 

We observed that SAM2 tends to over-segment images in the RWTD dataset, as can be seen in Figure~\ref{fig:teaser}. 
% Indeed, Figure \ref{fig:masks-count} compares the number of predicted segments of SAM2 and textureSAM on the RWTD. 
% As can be seen, SAM tends to over-segment images, whereas TextureSAM produces fewer and more coherent segments that better align with the underlying texture.
Therefore, in the following experiments we add a 
mask aggregation step, on top of SAM2, to evaluate the model’s ability to capture texture-based regions without excessive fragmentation that stems from shape bias. 

\begin{figure}[t]  
    \centering 

    \includegraphics[width=1.0\linewidth]{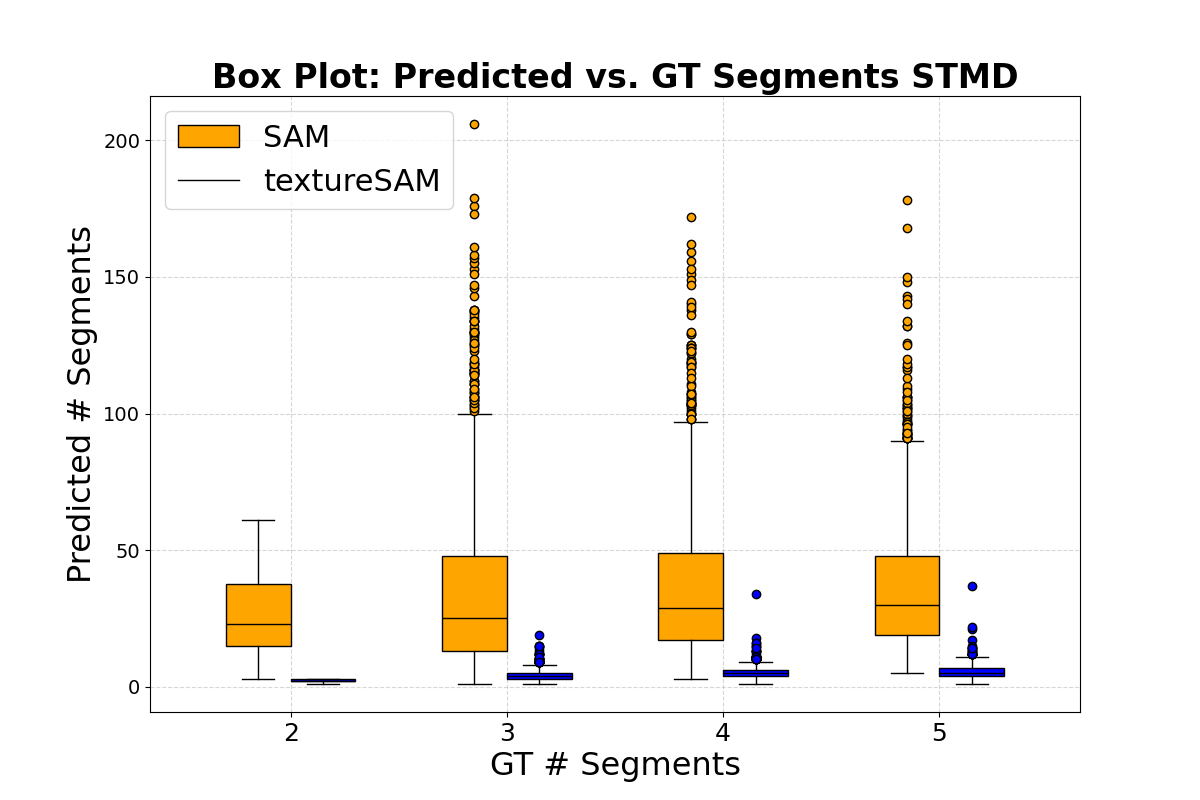}
    \caption{
    Box plot comparing predicted segments to the ground truth (GT) for the Synthetic Textured Masks Dataset (STMD). We group the results by the number of GT segments per image. SAM2 fragmentation of textures can be seen in the plot as it generates significantly more masks (segments).
    }
    \label{fig:masks-size}

\end{figure}

The synthetic dataset serves as an additional validation, allowing us to test the generalization capability of our approach in a controlled setting where textured regions are clearly defined in the ground truth and no objects appear in the image. This is further illustrated in Figure \ref{fig:masks-size}, where we observe that SAM tends to fragment textured regions into multiple smaller segments, while TextureSAM captures entire textured areas more effectively. Below, we detail our findings across these datasets and evaluation protocols.

%%%%%%%%%%%%%%%%

% \begin{figure}[t]  
%     \centering
%     \includegraphics[width=1.0\linewidth]{figs/boxplot_predicted_vs_gt_segments_natural.png}
%     \caption{
%     Box plot comparing predicted segments to the ground truth (GT) for the Real World Textured Dataset (RWTD). Each annotation contains two GT segments. SAM2’s over-segmentation is evident by it producing significantly more masks compared to TextureSAM.
%     }
%     \label{fig:masks-count-RWTD}
% \end{figure}

%

\subsection{Mask Aggregation Analysis}

To evaluate the overall segmentation quality of TextureSAM vs. the original SAM2 model, we apply mask aggregation, where predicted segmentation masks are grouped based on their overlap with ground truth regions. This approach provides a more holistic measure of segmentation performance by consolidating fragmented predictions that belong to the same texture-defined region. We wish to produce a model that recognizes regions with unique repeating patterns as a whole. Therefore, we also report metrics without aggregation, where individual predicted masks are evaluated independently. This penalizes models that over-segment regions by producing multiple small fragments instead of a single coherent mask for a textured region.

% Evaluation metrics include mean Intersection over Union (mIoU) which is the standard in assessing segmentation quality, as well as Adjusted Rand Index (ARI), that quantifies the similarity between predicted and GT masks.

% We use mean Intersection over Union (mIoU) to assess segmentation quality after aggregation, as it reflects how well the model identifies texture-based regions. Keeping in mind RQ1 (is SAM shape-biased?), both ARI and mIoU are reported on non-aggregated predictions to quantify segmentation fragmentation, which arises due to a tendency to over-segment textured regions. This over-segmentation is a direct consequence of shape bias, which favors segmenting object-like structures rather than grouping regions based on texture similarity. By reporting aggregated and non-aggregated results, we analyze both the segmentation accuracy and the degree of shape bias-induced fragmentation. The following sections present dataset-specific results while referring to this aggregation methodology.

% We apply mask aggregation in both the RWTD natural images dataset and the synthetic texture-shift dataset, where ground truth regions are defined by texture rather than object boundaries. The following sections present dataset-specific results while referring to this aggregation methodology.

\subsection{Synthetic STMD Dataset Results}

The synthetic texture-shift dataset provides a controlled benchmark where segmentation is based purely on texture, with no object structures present. This allows us to isolate the effect of texture on segmentation performance without interference from shape-based cues. As shown in Table~\ref{tab:synthetic_results}, TextureSAM with $\eta \leq 1.0$ achieves the best mIoU and ARI scores, demonstrating improved segmentation performance. The $\eta \leq 1.0$ training includes images that are defined solely by texture boarders, similar to the images in the STMD, making this result logical.
TextureSAM outperforms SAM-2 in both mIoU and ARI, demonstrating improved alignment with texture-defined ground truth regions. 

Applying mask aggregation further reveals the impact of shape bias in the original SAM-2 as it performs poorly. Interestingly, our inference parameters also benefit the original SAM-2 in this scenario, as it struggles to obtain confident masks from semantic-less data. With aggregation, both models achieve higher mIoU, as fragmented predictions are consolidated into coherent regions. However, here TextureSAM achieves only comparable results to the original SAM-2 with the modified parameters indicating our model retains overall segmentation capability, while being texture aware. Without aggregation, the original SAM-2 exhibits significantly lower ARI and mIoU, confirming that it tends to over-segment textured regions into multiple smaller components due to its inherent preference for shape-based segmentation. In contrast, TextureSAM mitigates this fragmentation, leading to more consistent segmentations that adhere to texture boundaries rather than arbitrary shape structures.

%%%%%%%%%%%%%%

\begin{table}[b]
\begin{center}
\begin{tabular}{|c|c|c|c|}
\hline
\textbf{STMD Results} & \textbf{mIoU} & \textbf{ARI} & \textbf{mIoU, Aggr.} \\
\hline
SAM-2 & 0.07 & 0.16 & 0.16 \\
SAM-2* & 0.17 & 0.16 & \textbf{0.78} \\
TextureSAM $\eta \leq 0.3$ & 0.33 & 0.32 & 0.71 \\
TextureSAM $\eta \leq 1.0$ & \textbf{0.35} & \textbf{0.34} & 0.70 \\
\hline
\end{tabular}
\end{center}
\caption{Results for the STMD synthetic image dataset. SAM-2* indicates using SAM-2 with the parameters used for TextureSAM. TextureSAM trained with strong texture augmentations attained the highest mIoU and ARI scores, outperforming both the original SAM-2 and the mild augmentations ($\eta \leq 0.3$) version of TextureSAM. For aggregated masks (mIoU, Aggr.), the original SAM-2 with its original inference parameters has the lowest mIoU, but interestingly, when using TextureSAM's inference parameters, the original SAM-2's score is the highest. This suggests that for synthetic images containing no salient objects, SAM-2 struggles to obtain high-confidence predictions, leading to sparse coverage of the GT-defined area.}
\label{tab:synthetic_results}
\end{table}

%%%%%%%%%%%%%%%%%%%%%%%%%%%

% \begin{table*}
% \begin{center}
% \begin{tabular}{|c|c|c|c|c|}
% \hline
%  \multicolumn{1}{|c|}{STMD Results} &
%  \multicolumn{1}{c|}{Original SAM DF} &
%  \multicolumn{1}{c|}{Original SAM} &
%  \multicolumn{1}{c|}{TextureSAM $\eta \ge 0.3$} &
%  \multicolumn{1}{c|}{TextureSAM $\eta \ge 1.0$} \\
% \hline\hline
% mIoU & $0.077$ & $0.1668$ & $0.3302$ & \textbf{0.3515 (+0.1847)} \\
% ARI & $0.1673$ & $0.1635$ & $0.3223$ & \textbf{0.3462 (+0.1789)} \\
% mIoU, aggregated masks & $0.1646$ & \textbf{0.785} & $0.712$ & $0.696$ \\
% \hline
% \end{tabular}
% \end{center}
% \caption{Results for the STMD synthetic image dataset. DF indicates using SAM's default inference parameters. TextureSAM trained with strong texture augmentations attained the highest mIoU and ARI scores, outperforming both the original SAM and the mild augmentations ($\eta \ge 0.3$) version of TextureSAM. For the aggregated masks, mIoU of the original SAM with it's original inference parameters is the lowest, but interestingly when using TextureSAM's inference parameters the original SAM's score is the highest. This indicates that for synthetic images containing no salient objects, SAM struggles with obtaining high confidence predictions. These predictions cover the GT defined area but are massively fragmented.}
% \label{tab:synthetic_results}
% \end{table*}

\begin{figure}[t]
\centering
\setlength{\tabcolsep}{1pt} % Adjust spacing between columns
\renewcommand{\arraystretch}{1.0} % Adjust row spacing
\begin{tabular}{c c c c c}
    {\raisebox{3\height}{\small \textbf{Image}}} &  
    \includegraphics[width=0.18\linewidth]{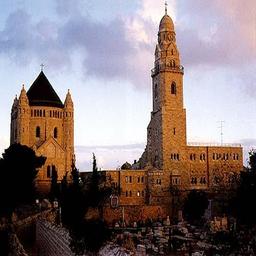} &
    \includegraphics[width=0.18\linewidth]{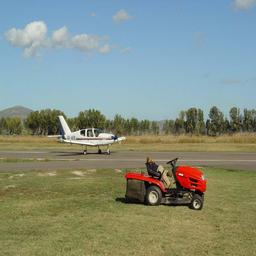} &
    \includegraphics[width=0.18\linewidth]{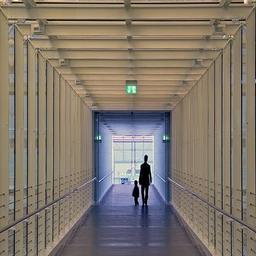} &
    \includegraphics[width=0.18\linewidth]{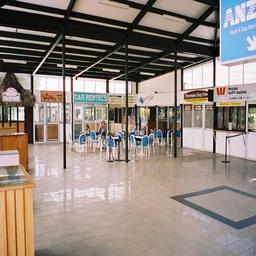} \\

    {\raisebox{3\height}{\small \textbf{SAM-2}}} &  
    \includegraphics[width=0.18\linewidth]{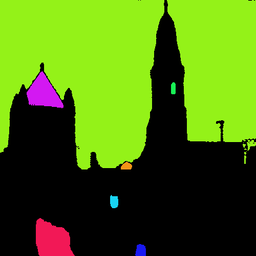} &
    \includegraphics[width=0.18\linewidth]{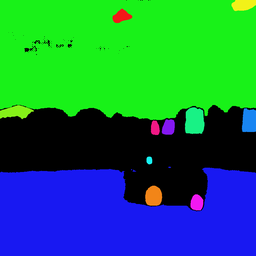} &
    \includegraphics[width=0.18\linewidth]{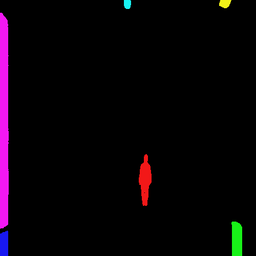} &
    \includegraphics[width=0.18\linewidth]{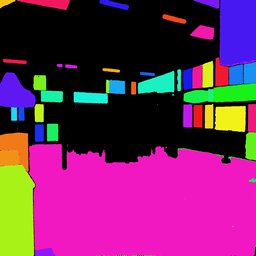} \\

    {\raisebox{3\height}{\small \textbf{SAM-2*}}} &  
    \includegraphics[width=0.18\linewidth]{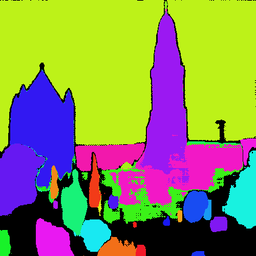} &
    \includegraphics[width=0.18\linewidth]{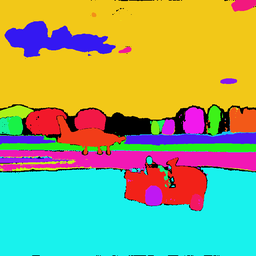} &
    \includegraphics[width=0.18\linewidth]{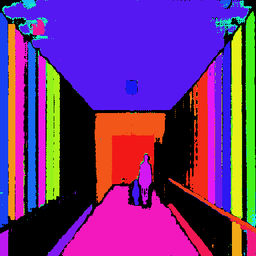} &
    \includegraphics[width=0.18\linewidth]{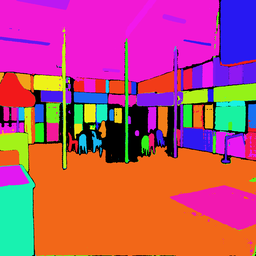} \\

    {\raisebox{3\height}{\small \textbf{TextureSAM}}} &  
    \includegraphics[width=0.18\linewidth]{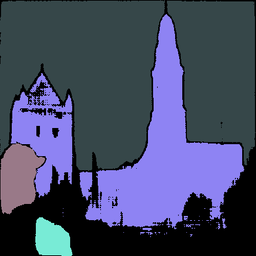} &
    \includegraphics[width=0.18\linewidth]{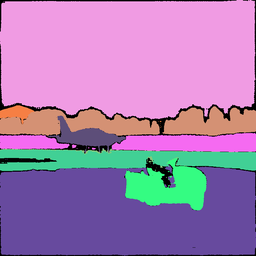} &
    \includegraphics[width=0.18\linewidth]{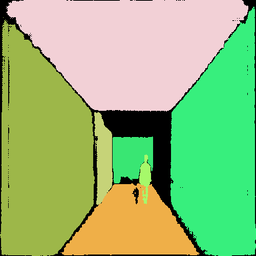} &
    \includegraphics[width=0.18\linewidth]{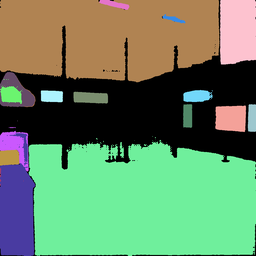} \\

    {\raisebox{3\height}{\small \textbf{GT}}} &  
    \includegraphics[width=0.18\linewidth]{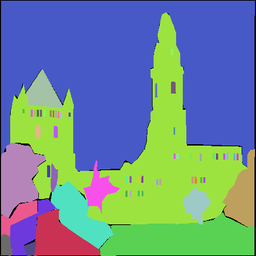} &
    \includegraphics[width=0.18\linewidth]{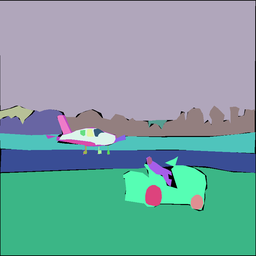} &
    \includegraphics[width=0.18\linewidth]{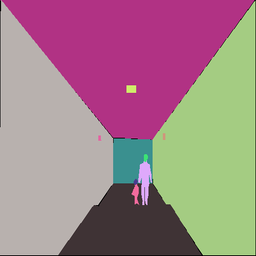} &
    \includegraphics[width=0.18\linewidth]{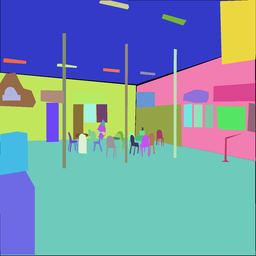} \\
\end{tabular}

\caption{Segmentation results on images  from the ADE20K dataset (1st row). It can be seen that TextureSAM (3rd row) produces comparable semantic segmentation to the original SAM-2 (2nd row, 3rd row with modified inference parameters). TextureSAM's predictions align better with the GT, where entire textured regions (e.g. trees, walls.) are marked with the same instance.}
\label{fig:ADE20K_results}
\end{figure}

\begin{figure}[t]
\centering
\setlength{\tabcolsep}{1pt}
\renewcommand{\arraystretch}{1.0} 
\begin{tabular}{c c c c c}
    {\raisebox{3\height}{\small \textbf{Image}}} &  
    \includegraphics[width=0.18\linewidth]{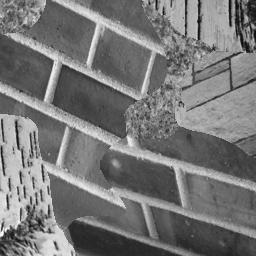} &
    \includegraphics[width=0.18\linewidth]{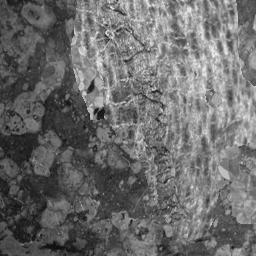} &
    \includegraphics[width=0.18\linewidth]{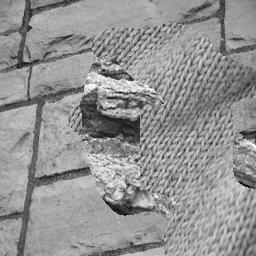} &
    \includegraphics[width=0.18\linewidth]{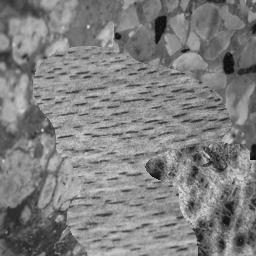} \\

    {\raisebox{3\height}{\small \textbf{SAM-2}}} &  
    \includegraphics[width=0.18\linewidth]{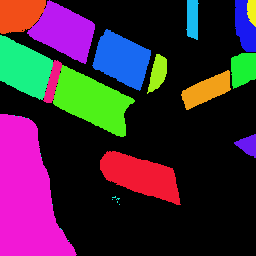} &
    \includegraphics[width=0.18\linewidth]{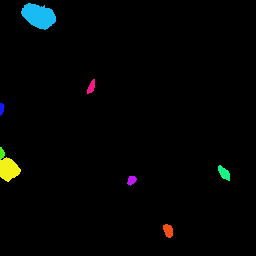} &
    \includegraphics[width=0.18\linewidth]{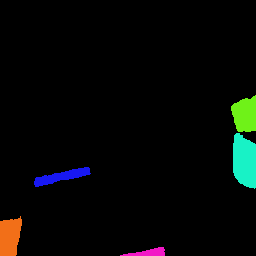} &
    \includegraphics[width=0.18\linewidth]{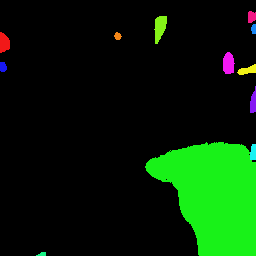} \\

    {\raisebox{3\height}{\small \textbf{SAM-2*}}} &  
    \includegraphics[width=0.18\linewidth]{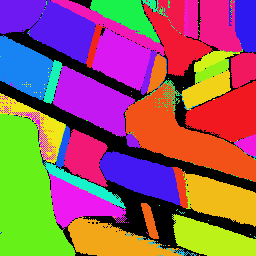} &
    \includegraphics[width=0.18\linewidth]{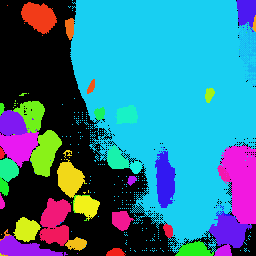} &
    \includegraphics[width=0.18\linewidth]{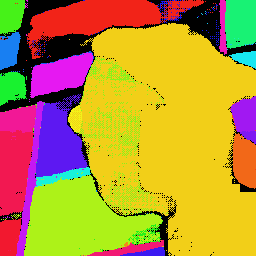} &
    \includegraphics[width=0.18\linewidth]{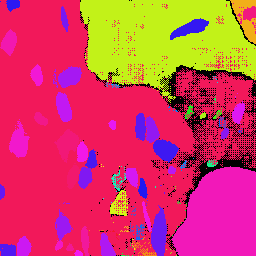} \\

    {\raisebox{3\height}{\small \textbf{TextureSAM}}} &  
    \includegraphics[width=0.18\linewidth]{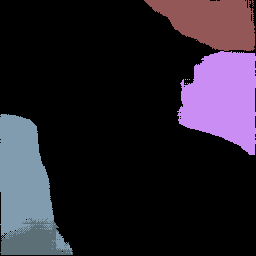} &
    \includegraphics[width=0.18\linewidth]{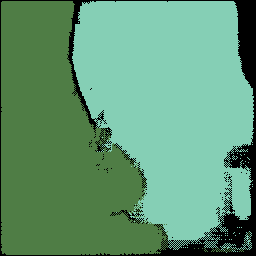} &
    \includegraphics[width=0.18\linewidth]{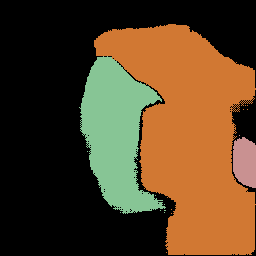} &
    \includegraphics[width=0.18\linewidth]{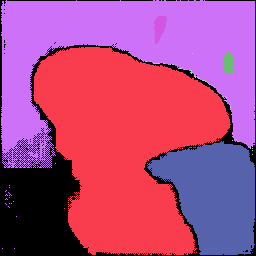} \\

    {\raisebox{3\height}{\small \textbf{GT}}} &  
    \includegraphics[width=0.18\linewidth]{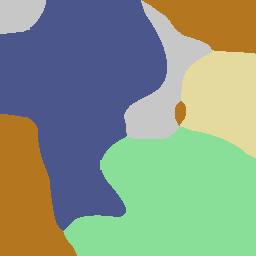} &
    \includegraphics[width=0.18\linewidth]{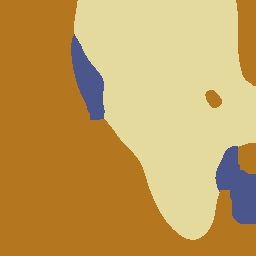} &
    \includegraphics[width=0.18\linewidth]{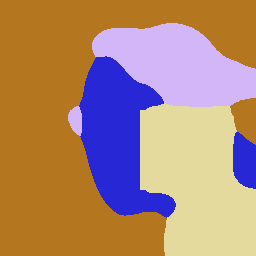} &
    \includegraphics[width=0.18\linewidth]{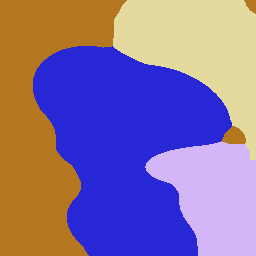} \\
\end{tabular}
\caption{Segmentation results on the synthetic STMD dataset. 1st row shows original images, the following rows present segmentation by the different models and the GT annotations. For this semantic-less dataset, TextureSAM segmentation maps better align with GT annotations, while SAM-2 fragments textured regions into individual elements.}
\label{fig:synthetic_results}
\end{figure}

\subsection{Real-World Segmentation Results}

On the Real-World Textured Dataset (RWTD), which comprises natural images, we observe a consistent improvement in segmentation performance with TextureSAM compared to SAM-2. As RWTD is specifically designed for texture-based segmentation, this dataset provides a strong benchmark for evaluating how well models can capture texture-defined regions rather than relying on shape cues. Figure~\ref{fig:teaser} provides a qualitative visualization of the difference between the original SAM-2 model and TextureSAM. SAM-2 tends to fragment textured regions based on semantic elements comprising the texture. Further box-plot visualization of the number of masks predicted by SAM-2 is presented in the supplemental material, and is similar to that presented for the STMD dataset. Quantitative results are presented in Table~\ref{tab:natural_results}, where TextureSAM with $\eta \leq 0.3$ achieves the best mIoU and ARI scores, demonstrating improved texture-aware segmentation performance.

When applying mask aggregation, TextureSAM obtains comparable, and slightly higher results to the original SAM-2. However, without aggregation, SAM-2 exhibits a notable drop in ARI and mIoU, confirming its tendency to over-segment texture-defined regions due to shape bias. When comparing the performance of the original SAM-2 with the different inference parameter we see a notable drop in segmentation ability of natural images when using our modified inference parameters.

%%%%%%%%%%%%%%%%%%%%%%%%%

\begin{table}
\begin{center}
\begin{tabular}{|c|c|c|c|}
\hline
\textbf{RWTD Results} & \textbf{mIoU} & \textbf{ARI} & \textbf{mIoU, Aggr.} \\
\hline
SAM-2 & 0.26 & 0.36 & 0.44 \\
SAM-2* & 0.14 & 0.19 & 0.75 \\
TextureSAM $\eta \leq 0.3$ &  \textbf{0.47} &  \textbf{0.62} & 0.75 \\
TextureSAM $\eta \leq 1.0$ & 0.42 & 0.54 &  \textbf{0.76} \\
\hline
\end{tabular}
\end{center}
\caption{Results for the RWTD natural image dataset. SAM-2* indicates using SAM-2 with the parameters used for TextureSAM. TextureSAM trained with mild texture augmentations ($\eta \leq 0.3$) attained the highest mIoU and ARI scores, significantly surpassing the original SAM-2. For aggregated masks (mIoU, Aggr.), both TextureSAM variants outperform the original SAM-2 with its default parameters. Using TextureSAM’s inference parameters on the original SAM-2 produces comparable results. Similar to STMD results, this suggests SAM-2 struggles with confident predictions in texture-driven images, reinforcing its shape bias.}
\label{tab:natural_results}
\end{table}

%%%%%%%%%%%%%%%%%%%%%%%%%

% \begin{table*}
% \begin{center}
% \begin{tabular}{|c|c|c|c|c|}
% \hline
%  \multicolumn{1}{|c|}{RWTD Results} &
%  \multicolumn{1}{c|}{\shortstack{Original SAM DF}} &
%  \multicolumn{1}{c|}{Original SAM} &
%  \multicolumn{1}{c|}{TextureSAM $\eta \ge 0.3$} &
%  \multicolumn{1}{c|}{TextureSAM $\eta \ge 1.0$} \\
% \hline\hline
% mIoU & $0.2641$ & $0.1468$ & \textbf{0.468 (+0.2039)} & $0.4293$ \\
% ARI & $0.3628$ & $0.1917$ & \textbf{0.616 (+0.2532)} & $0.545$ \\
% mIoU, aggregated masks & $0.441$ & $0.757$ & $0.752$ & \textbf{0.769 (+0.012)} \\
% \hline
% \end{tabular}
% \end{center}
% \caption{Results for the RWTD natural image dataset. DF indicates using SAM's default inference parameters. TextureSAM trained with only mild texture augmentations  ($\eta \ge 0.3$) attained the highest mIoU and ARI scores, significantly surpassing the original SAM. For the aggregated masks, mIoU of both TextureSAM's variants outperform the original SAM with it's default inference parameters. When using TextureSAM's inference parameters with the original SAM we obtain comparable and slightly better results. Similarly to the STMD dataset results, this indicates that for natural images containing no salient objects, SAM struggles with obtaining high confidence predictions, thus reaffirming it's shape-bias.}
% \label{tab:natural_results}
% \end{table*}

\subsection{ADE20K Validation Dataset Segmentation Results}

Finetuning SAM-2 to be more texture-aware comes at the risk of damaging it's existing semantic segmentation capabilities. We use ADE20k in an attempt to avoid catastrophic-forgetting, as it is a part of the original dataset used to train SAM-2. ADE20k's validation dataset is a challenging semantic segmentation benchmark, even for the SAM-2 original model. The validation dataset is therefore useful  to ascertain TextureSAM's overall semantic segmentation capabilities vs. the original SAM-2 model. Table \ref{tab:ADE20k_results} shows a comparison of the original SAM-2, with both default and TextureSAMs inference parameters, as well as TextureSAM mild augmentation model ($\eta \leq 0.3$) and severe augmentation ($\eta \leq 1.0$). It can be observed that both varients of TextureSAM achieve higher scores in ARI and mIoU  compared to the original SAM-2 model, further indicating that SAM-2 over-fragmentes GT areas. The Original SAM-2 with the modified inference parameters scores the highest in the aggregated mIoU score, which is the most relevant for semantic segmentation evaluation. TextureSAM with mild augmentations $\eta \leq 0.3$ achieves a comparable (-0.1 mIoU) score indicating that shifting the model's focus towards texture did reduce semantic segmentation capabilities to a degree. Using the modified inference parameters increased the original SAM-2's score by producing more masks. This increase was also recorded for the texture datasets, where TextureSAM still proved superior. Figure~\ref{fig:ADE20K_results} further illustrates the visual differences between SAM-2 and TextureSAM in ADE20K, showing how TextureSAM provides more coherent segmentation of texture-defined regions compared to SAM-2.

%For the aggregated masks analysis the $\eta \ge 1.0$ TextureSAM achieves the highest score, with the $\eta \ge 0.3$ TextureSAM attaining comparable scores to the original SAM model. We therefore conclude that TextureSAM's semantic segmentation capabilities survive our texture-awareness finetuning protocol with a substantial benefit on the ADE20k validation dataset. 

%%%%%%%%%%%%%%%

\begin{table}[t]
\begin{center}
\begin{tabular}{|c|c|c|c|}
\hline
\textbf{ADE20k Results}  & \textbf{mIoU} & \textbf{ARI} & \textbf{mIoU,Aggr.} \\
\hline
SAM-2  & 0.08 & 0.08 & 0.46 \\
SAM-2*  & 0.1 & 0.11 & \textbf{0.65} \\
TextureSAM $\eta \leq 0.3$  & 0.11 & 0.11 & 0.55 \\
TextureSAM $\eta \leq 1.0$  & 0.12 & 0.11 & 0.40 \\
\hline
\end{tabular}
\end{center}
\caption{Results for the ADE20k natural image dataset. SAM-2* indicates using SAM-2 with the parameters used for TextureSAM. SAM-2* scores the highest in the aggregated mIoU score (mIoU,Aggr.), which is the most relevant for semantic segmentation evaluation. TextureSAM with mild augmentations $\eta \leq 0.3$ achieves a comparable (-0.1 mIoU) score indicating that shifting the model's focus towards texture did reduce semantic segmentation capabilities to a degree. Using the modified inference parameters increased the original SAM-2's score by producing more masks. This increase was also recorded for the texture datasets, where TextureSAM still proved superior.}
\label{tab:ADE20k_results}
\end{table}

\section{Conclusions}

In this work we have demonstrated SAM-2's shape-bias, via evaluation on texture-centric datasets. This provides strong empirical evidence that agree with previous works. Therefore, in regard to RQ1, based on our findings, SAM-2 is indeed shape-biased. There is some evidence in our work that suggest that SAM-2 is able to comprehend textures as a whole regions, yet it is strongly coupled with shape as evident by more frequent fragmentation of textured regions.

To address this issue, we introduce TextureSAM, a fine-tuned variant of SAM, that incorporates targeted texture augmentations during training. TextureSAM successfully mitigates SAM’s shape bias, leading to improved segmentation in texture-driven scenarios. 

In regard to RQ2, we provide empirical evidence that it is indeed possible to shift a foundation model towards performing texture-driven segmentations. These observations highlight a trade-off between enhancing texture sensitivity and preserving broader segmentation performance, pointing to the need for careful calibration of texture-focused training protocols.

%%%%%%%%%%%%%%%%%%%%%

These findings underscore the complex interplay between shape and texture in segmentation tasks, offering deeper insight into how fine-tuning strategies can shift a model’s focus. SAM’s reliance on high-level semantics often causes it to overlook subtle textural cues, whereas TextureSAM—especially under moderate augmentation—more reliably identifies texture boundaries. However, the performance gains come with notable trade-offs: heavy augmentation can yield advantages for synthetic or uniformly textured domains (e.g., STMD) but deviates too far from natural image distributions, hindering performance on real-world data. Conversely, moderate augmentation strikes a middle ground by enhancing texture sensitivity without sacrificing alignment with typical image statistics. This tension highlights how different applications may require different balances of shape and texture emphasis, from industrial inspection (where minor texture changes are critical) to more conventional object-focused tasks.

\section*{Acknowledgments}
This work was supported by the Pazy Foundation.

{
    \small
    \bibliographystyle{ieeenat_fullname}
    \bibliography{egbib}
}

%\clearpage
%\onecolumn
%\input{arxiv_app}

\end{document}